\documentclass[runningheads]{llncs}

\usepackage{eccv}

\usepackage{eccvabbrv}

\usepackage{graphicx}
\usepackage{booktabs}

\usepackage{epsfig}
\usepackage{amsmath}
\usepackage{amssymb}
\usepackage{multirow}
\usepackage{colortbl}
\definecolor{lightgreen}{rgb}{0.75, 0.95, 0.69}

\usepackage[accsupp]{axessibility}  

\usepackage{hyperref}

\usepackage{orcidlink}

\begin{document}
\title{Masked Angle-Aware Autoencoder for Remote Sensing Images} 

\titlerunning{Masked Angle-Aware Autoencoder}

\author{Zhihao Li\orcidlink{0000-0001-7119-3215} \and
Biao Hou$^\dag$ \orcidlink{0000-0002-1996-186X} \and
Siteng Ma\orcidlink{0000-0001-9678-0213} \and
Zitong Wu\orcidlink{0000-0002-0449-9465} \and 
Xianpeng Guo\orcidlink{0000-0003-3733-2570} \and
Bo Ren\orcidlink{0000-0002-0481-5069} \and
Licheng Jiao\orcidlink{0000-0003-3354-9617}}
\authorrunning{Z. Li et al.}

\institute{School of Artificial Intelligence, Xidian University \\
\email{avcodec@163.com} \\
$^\dag$ Corresponding Author
}

\maketitle

\begin{abstract}
To overcome the inherent domain gap between remote sensing (RS) images and natural images, some self-supervised representation learning methods have made promising progress. However, they have overlooked the diverse angles present in RS objects. This paper proposes the Masked Angle-Aware Autoencoder (MA3E) to perceive and learn angles during pre-training. We design a \textit{scaling center crop} operation to create the rotated crop with random orientation on each original image, introducing the explicit angle variation. MA3E inputs this composite image while reconstruct the original image, aiming to effectively learn rotation-invariant representations by restoring the angle variation introduced on the rotated crop. To avoid biases caused by directly reconstructing the rotated crop, we propose an Optimal Transport (OT) loss that automatically assigns similar original image patches to each rotated crop patch for reconstruction. MA3E\footnote{Our code will be released at: \url{https://github.com/benesakitam/MA3E}} demonstrates more competitive performance than existing pre-training methods on seven different RS image datasets in three downstream tasks.
\end{abstract}

\section{Introduction}
Nowadays, deep learning-based interpretation of remote sensing (RS) images has been widely applied in fields related to national defense~\cite{zhang2020domain, schumann2018assisting, rolnick2022tackling} and people's well-being~\cite{gu2019survey, mulla2013twenty, ippoliti2019defining}. The increasing number of Earth observation satellites makes it possible to acquire a massive amount of unlabeled RS images. Despite the abundance of data, many RS models still initialize with the ImageNet~\cite{deng2009imagenet} pre-trained weights. Inherent domain gap between natural images and RS images limit the performance of these models. Therefore, exploring self-supervised representation learning on RS images is highly necessary.

\begin{figure}[tb]
\centering
  \begin{subfigure}{0.49\linewidth}
    \includegraphics[width=1\textwidth]{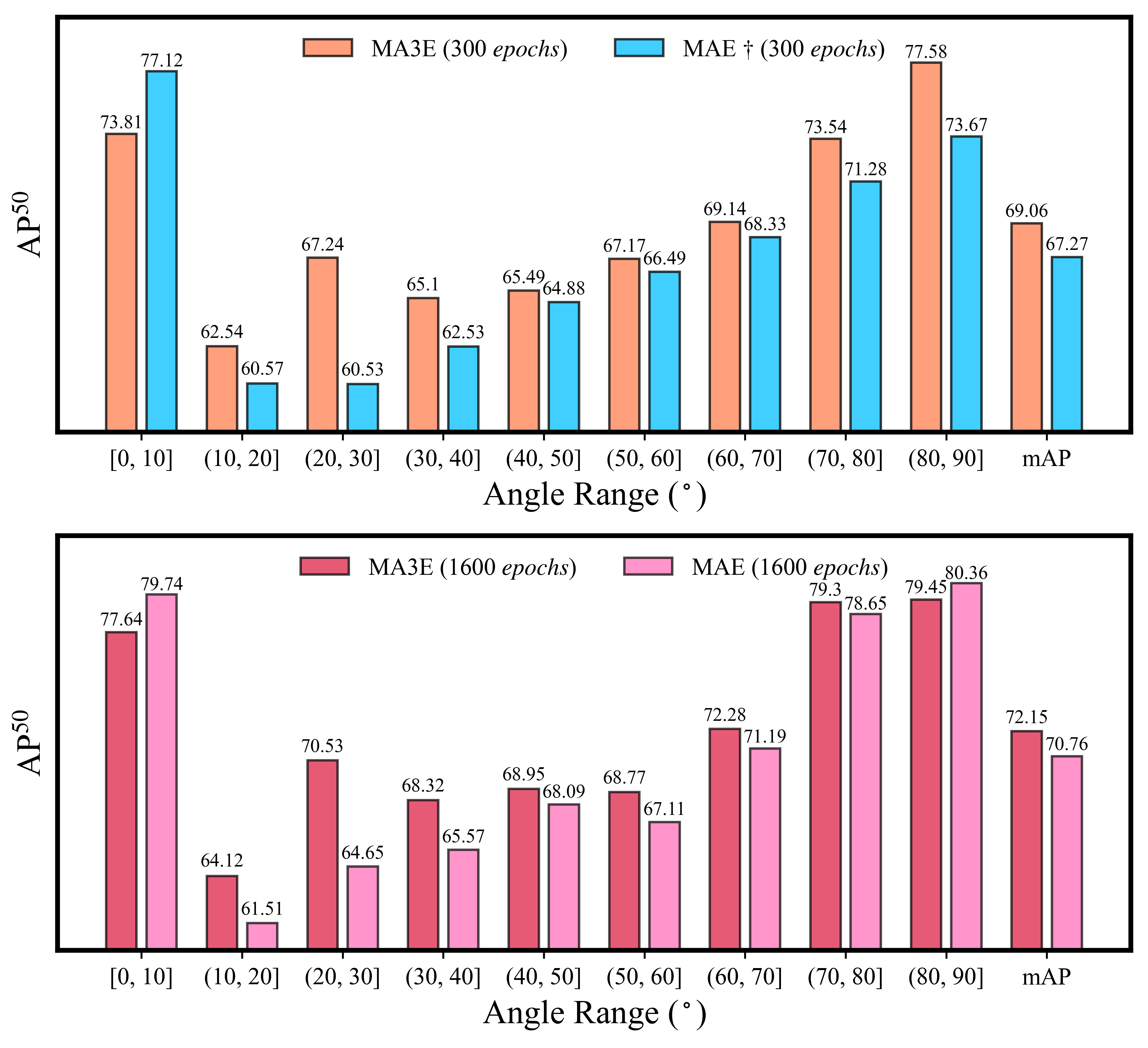}
    \caption{Detection results on DOTA1.0~\cite{xia2018dota}. Due to labels of the testing set is unavailable, we train on the training set and report AP$^{50}$ on the validation set.}
    \label{fig1-a}
  \end{subfigure}
  \hfill
  \begin{subfigure}{0.49\linewidth}
    \includegraphics[width=1\textwidth]{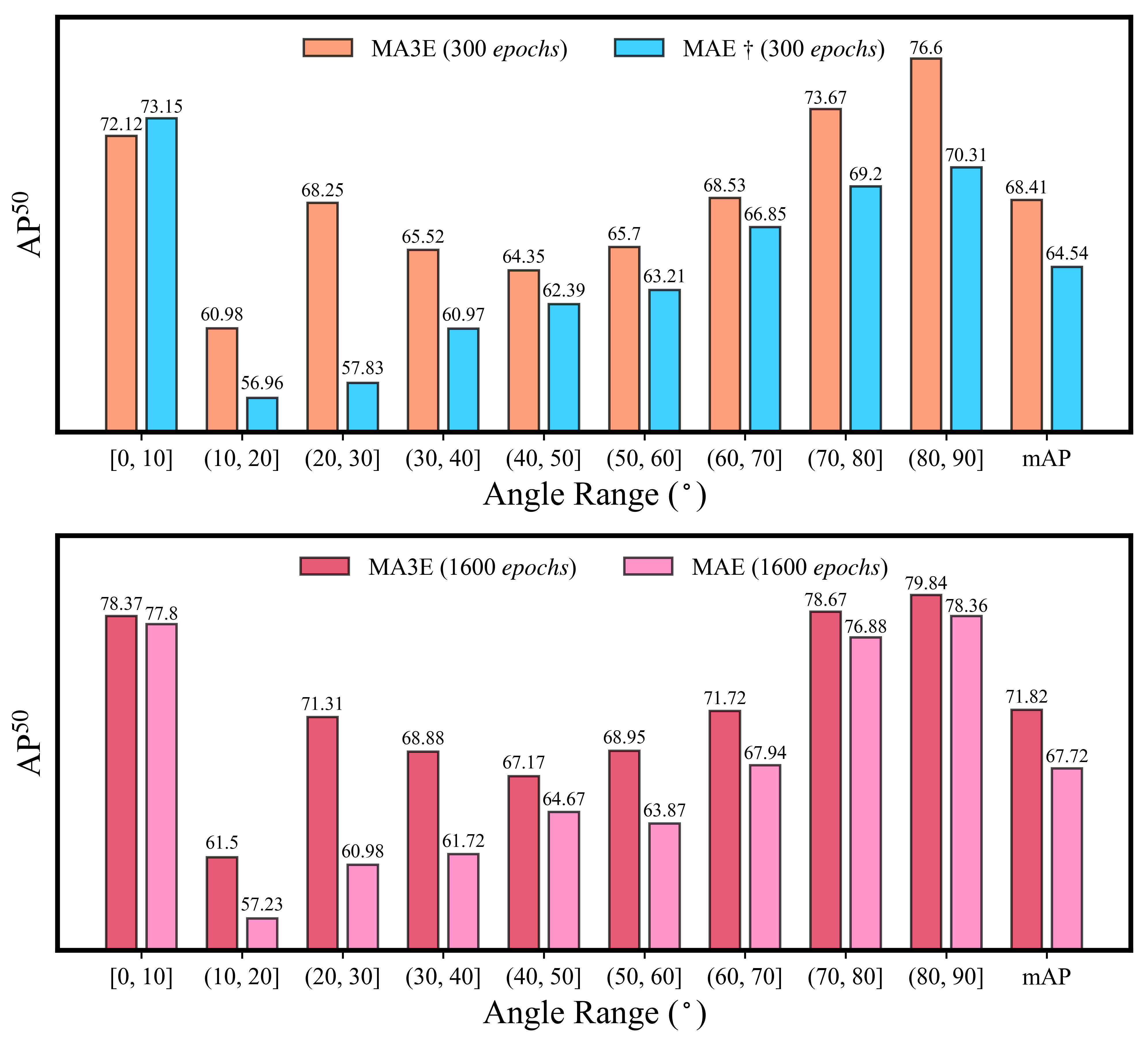}
    \caption{Detection results on DIOR-R~\cite{cheng2022anchor}. We train on the training set and report AP$^{50}$ on the testing set. ~\\}
    \label{fig1-b}
  \end{subfigure}
\caption{Detection results of the detector loaded with MA3E and MAE~\cite{wang2022advancing} pre-trained models for RS objects categorized into different angle ranges. The fine-tuning experimental setup is the same as described in Sec.~\ref{sec4.1}. Our MA3E models, pre-trained for 300 \textit{epochs} and 1600 \textit{epochs}, notably enhance AP$^{50}$ for objects with angles ranging from 10$^\circ$ to 80$^\circ$, demonstrating the effectiveness of angle perception during pre-training. \dag \ denotes our reproduction, as \emph{Wang} \textit{et al.}~\cite{wang2022advancing} only releases the model pre-trained for 1600 \textit{epochs} using MAE~\cite{he2022masked} on an RS image dataset.}
\label{fig1}
\end{figure}

Self-supervised representation learning~\cite{chen2020simple, he2020momentum, bao2021beit, he2022masked} for natural images has emerged as a new paradigm for pre-training models on large-scale datasets. Among these, Masked Image Modeling (MIM)~\cite{bao2021beit, he2022masked, xie2022simmim, wei2022masked} learns visual representations by reconstructing masked portions of the input. With its concise architecture and outstanding performance on downstream tasks~\cite{deng2009imagenet, lin2014microsoft, zhou2019semantic}, it has attracted widespread attention. Recently, several noteworthy MIM studies have surfaced in the RS image field~\cite{cong2022satmae, reed2023scale, wang2022advancing}, offering excellent initialization for the vision transformer~\cite{dosovitskiy2020image, zhang2023vitaev2} and achieving good results across various downstream tasks. This demonstrates the potential of MIM in representation learning for RS images.

Although existing customized MIM methods for RS images take various factors into account, such as different resolutions~\cite{reed2023scale}, multi-scale objects and the complex background~\cite{sun2022ringmo}, and imaging from multiple spectral bands~\cite{cong2022satmae}, they are not effective learners in the face of angles of RS objects. Objects in natural images typically have fixed orientations due to gravity, whereas in RS images, objects captured from an overhead perspective often exhibit a wide range of angles. The same RS object presents diverse shapes and appearances when viewed from different angles. Properly perceiving and considering angle information aligns with the nature of how objects are captured in the RS community, which is conducive to accurate image interpretation. The above methods only focus on reconstructing the pixel values of RS objects, the learning for angles is often implicitly accompanied within reconstruction. Therefore, we propose the {\bf M}asked {\bf A}ngle-{\bf A}ware {\bf A}uto{\bf e}ncoder (MA3E), which perceives and learns angle information by restoring the preset angle variations during original pixel reconstruction. As an illustrative example, when fine-tuning a pre-trained model for rotated object detection, we count the angles of all correctly detected objects in two datasets and report AP$^{50}$ for objects falling within different angle ranges in Fig.~\ref{fig1}. \cite{wang2022advancing} only obtains the higher AP$^{50}$ for objects with angles close to horizontal (\textit{e.g.}, 0$^\circ$ or 90$^\circ$). In contrast, MA3E significantly improves AP$^{50}$ for objects with large inclinations. This indicates that MA3E effectively becomes aware of diverse angles of objects and learn robust rotation-invariant representations.

MA3E follows an asymmetric encoder-decoder architecture and the similar pre-training principle of the classical Masked Autoencoder (MAE)~\cite{he2022masked}. Firstly, MA3E creates the rotated crop at arbitrary position within each RS image, introducing an explicit angle variation. We propose a \textit{scaling center crop} operation to construct diverse rotated crops while preserving primary scenes. Each rotated crop is center-rotated at a random angle and replaces the original scene. Then, MA3E takes this composite image as the training input and reconstructs the original image. Before feeding into the encoder, an additional angle embedding is added to the rotated crop to implicitly prompt the model to perceive the angle of this region. Masking is applied respectively to the rotated crop and the remaining background to avoid losing too much or all patches of rotated crop. Due to the scene offset in the rotated crop, direct reconstruction using the original image patches of corresponding positions would result in obvious biases. Therefore, the reconstruction for the rotated crop is treated as an Optimal Transport (OT) problem. We propose an OT loss that automatically allocates similar original image patches as reconstruction targets for each patch of the rotated crop.

By simultaneously restoring the angle variations while reconstructing the original pixels. MA3E exhibits an awareness of diverse angles, enabling it to effectively learn rotation-invariant visual representations. MA3E demonstrates outstanding performance in several downstream tasks, including scene classification on NWPU-RESISC45~\cite{cheng2017remote}, AID~\cite{xia2017aid}, and UC Merced~\cite{yang2010bag}, rotated object detection on DOTA1.0~\cite{xia2018dota} and DIOR-R~\cite{cheng2022anchor}, as well as semantic segmentation on iSAID~\cite{waqas2019isaid} and \href{https://www.isprs.org/education/benchmarks/UrbanSemLab/2d-sem-label-potsdam.aspx}{Potsdam}.

\section{Related Works}

The development of vision transformers~\cite{dosovitskiy2020image, liu2021swin, wang2021pyramid, zhang2023vitaev2} has advanced the masked image modeling~\cite{bao2021beit, he2022masked, gao2022convmae}. MIM has gradually replaced contrastive learning~\cite{he2020momentum, chen2020simple, grill2020bootstrap}, becoming the currently prominent pre-training paradigm in the computer vision field.

{\bf Model Image Modeling.} MIM aims to reconstruct masked parts using the visible input. BEiT~\cite{bao2021beit} masks 60\% of the image and relies on tokens extracted from these masked regions by dVAE~\cite{rolfe2016discrete} for reconstruction. SimMIM~\cite{xie2022simmim} encodes visible patches and mask tokens, directly predicting the original pixel values. MAE~\cite{he2022masked} improves reconstruction efficiency by feeding only visible patches into the encoder. GreenMIM~\cite{huang2022green} proposes an optimal grouping algorithm, deploying MAE on a hierarchical transformer~\cite{liu2021swin} by dividing each window into multiple groups. Some studies design diverse reconstruction targets, such as advanced CLIP~\cite{hou2022milan} or DINO features~\cite{gao2023mimic}, HOG features~\cite{wei2022masked}, frequencies~\cite{xie2022masked, liu2023pixmim}, and multi-level features~\cite{wang2023masked, liu2023improving}. Moreover, some works pay attention to process input images; \cite{tian2022beyond} recovers masked patches with five different learning targets, LoMaR~\cite{chen2022efficient} reconstructs multiple local regions of an image, and MixMAE~\cite{liu2023mixmae} takes a mixed image as input and simultaneously reconstructs multiple original images before mixing. These methods significantly advance self-supervised representation learning based on natural images.

{\bf MIM in RS Images.} Imaging sources for RS images are diverse, covering complex scenes with uneven scales and distributions of foreground objects or land cover. Current works have gradually transitioned from contrastive learning utilizing land cover information such as seasonal changes~\cite{manas2021seasonal} and temporal differences~\cite{mall2023change} to the customized MIM methods. \textit{Wang et al.}~\cite{wang2022advancing} pre-trains with MAE~\cite{he2022masked} on the MillionAID dataset~\cite{long2021creating} and fine-tunes by replacing the original transformer's global attention with rotated varied-size window attention in downstream tasks. CMID~\cite{muhtar2023cmid} introduces contrastive learning into the MIM branch to learn consistency. RingMo~\cite{sun2022ringmo} collects two millions images from satellite and aerial platforms and designs a patch incomplete masking strategy for reconstruction. GFM~\cite{mendieta2023towards} pre-trains on GeoPile, a constructed dataset with multiple sources, and continual learns valuable in-domain representations under the guidance of the ImageNet-22k models. SatMAE~\cite{cong2022satmae} encodes temporal and multi-spectral information in position embeddings to extend spatio-temporal relationships in the fMoW dataset~\cite{christie2018functional}. ScaleMAE~\cite{reed2023scale} leverages the inherent ground sample distance to reconstruct multi-scale resolution images. It is regrettable that the above methods do not explore angles during pre-training. In this paper, we propose MA3E, which simultaneously performs pixel reconstruction and angle restoration, thus perceiving angles and learning rotation-invariant representations.

\section{Method}

\subsection{Preliminary: MAE}

MAE~\cite{he2022masked} employs an asymmetric encoder-decoder architecture for efficient masked image modeling. An input image, $x \in {\mathbb R}^{H \times W \times C}$, is first reshaped into a series of non-overlapping image patches of size $p \times p$, denoted as ${x^p} \in {\mathbb R}^{N \times {p^2}C}$, where $N=HW/{p^2}$ is the number of patches. Then, ${x^p} = \{x_i^p|i = 1,2,...,N\}$ is linearly mapped into patch embeddings. MAE adds positional encoding information to these embeddings and randomly masks them at a certain ratio, such as 75\%. The masked patches, ${x^m} = \{x_i^m|i = 1,2,...,N^m\}$, are discarded, and only the remaining visible patches, ${x^v} = \{x_i^v|i = 1,2,...,N^v\}$, are fed into the encoder to extract latent features. These latent features, along with the shared and learnable mask tokens representing the substituted masked patches, constitute the input to the decoder, with positional embeddings also added. After obtaining the decoder output $\hat x = \{{\hat x_i}|i = 1,2,...,N\}$, MAE predicts only the pixel values of the masked patches, using the original image patches as the reconstruction targets. This is achieved by computing the mean squared error (MSE) loss:

\begin{equation}
\label{eq1}
{\cal L}_{MSE}(x^m,\hat x^m) = \left\| {x^m} - {\hat x}^m \right\|_2^2,
\end{equation}
where ${\hat x^m} \in {\mathbb R}^{{N^m} \times {p^2}C}$ denotes the output of the decoder for masked patches. The decoder is adopts for pre-training only, while the encoder is further fine-tuned for downstream tasks. This often results in the common practice of using a lightweight decoder and a complete transformer encoder. The proposed MA3E shares the similar principle with MAE, described next.

\subsection{Masked Angle-Aware Autoencoder (MA3E)}

\begin{figure}[t]
\centering
\includegraphics[width=4.8in]{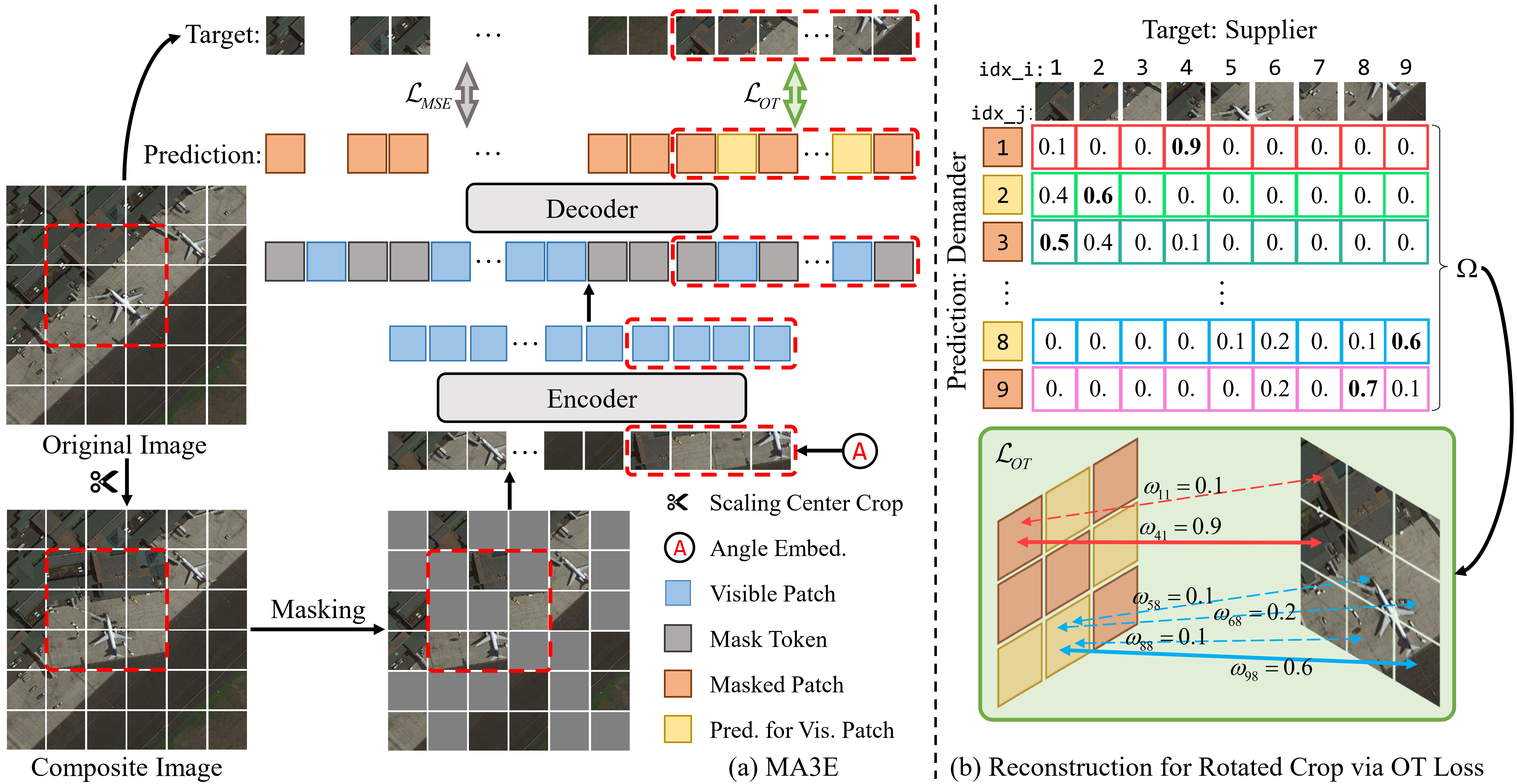}
\caption{(a) The pipeline of MA3E. A \textit{scaling center crop} operations is designed to create the rotated crop within the original image, introducing an explicit angle variation. An angle embedding is added to the rotated crop, followed by random masking the rotated crop along with the remaining background respectively. Then, all visible patches undergo sequential encoding and decoding to reconstruct the original image and restore the preset angle variation. (b) MA3E treats the reconstruction for rotated crops as an OT problem. By leveraging the Sinkhorn-Knopp fast iterative algorithm~\cite{cuturi2013sinkhorn} to solve the transportation plan $\Omega $, an OT loss is proposed. OT loss automatically assigns similar image patches for each predicted patch of the rotated crop for reconstruction.}
\label{fig2}
\end{figure}

MA3E aims to be aware of diverse angles and learn rotation-invariant visual representations. Fig.~\ref{fig2}(a) illustrates our pipeline. MA3E constructs rotated crops at arbitrary positions on the RS images by deploying the designed \textit{scaling center crop} operation to introduce explicit angle variations. These rotated crops have random angles and replace the scenes at their original locations. We add an angle embedding in each rotated crop, and mask the rotated crop and remaining background separately. For the reconstruction of the rotated crop, MA3E automatically allocates similar image patches as reconstruction targets for each rotated crop patch based on the transportation plan solved by the Sinkhorn-Knopp~\cite{cuturi2013sinkhorn} algorithm. This avoids biases introduced by the crop operation.

\begin{figure}[t]
\centering
\includegraphics[width=4.8in]{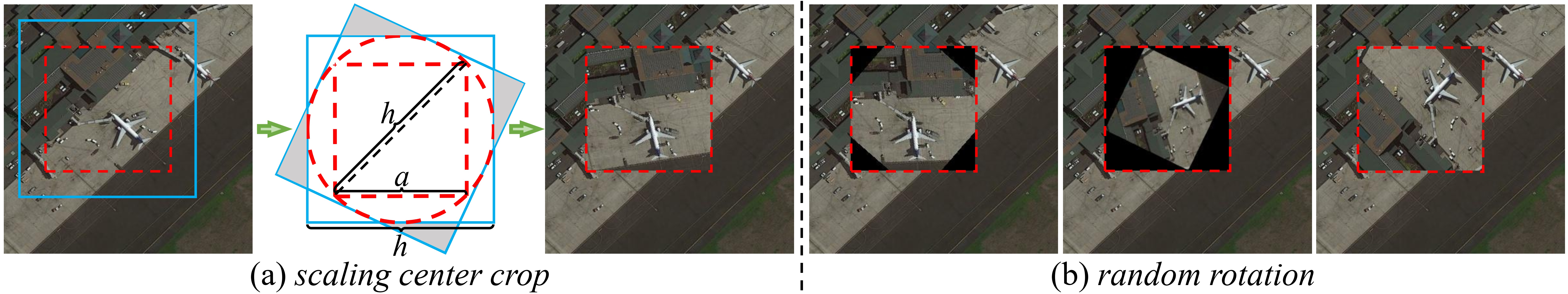}
\caption{(a) The proposed \textit{scaling center crop} constructs the rotated crop with a random angle at arbitrary position in the original image, introducing the explicit angle variation. (b) In the left and middle columns, the simple \textit{random rotation} operation results in i) meaningless background with zero values; ii) loss of the scene; iii) changes in scene scale. In the right column, fixed angles (\textit{e.g.}, 90°, 180°, 270°) for rotation restrict the diversity of scenes.}
\label{fig3}
\end{figure}

{\bf Rotated crop.} Using a simple \textit{random rotation} operation to construct the rotated crop of side length $a$ for each RS image would lead to adverse results as shown in Fig.~\ref{fig3}(b). The model struggles to learn high-quality representations from these regions, resulting in wasted computational resources. Therefore, we propose a \textit{scaling center crop} operation to create diverse rotated crops that preserve scenes to a substantial extent, as shown in Fig.~\ref{fig3}(a). For a square region (blue) of side length $h$ at arbitrary positions on an image, rotating this region at random angle would lead to loss of edge scenes (gray). However, the scenes within its largest inscribed circle (red) are fully preserved. Hence, we perform center cropping to extract the largest inscribed square region with side length $a = \frac{\sqrt 2}{2}h$ from the red circle as the rotated crop. This region holds an arbitrary orientation, replacing the original scene and introducing the explicit angle variation to the composite image. To ensure that each rotated crop can be entirely patchified, the side length $a$ needs to be divisible by the patch size $p$, with the starting position being a multiple of $p$.

{\bf Angle embedding.} In addition to adding positional embeddings for the divided patches of composite images, MA3E also includes the angle embeddings for the rotated crops. Each angle embedding is a learnable vector shared across each patch within a rotated crop. They serve as implicit cues for the model to perceive the angle variation in the rotated crop, while also distinguishing them from the remaining background.

{\bf Random masking.} Given $N_r = {a^2}/{p^2}$ patches from the rotated crop (denoted as $r = \{{r_i}|i = 1,2,...,{N_r}\}$) and $N_b = N - {N_r}$ patches from the background (denoted as $b = \{{b_i}|i = 1,2,...,{N_b}\}$), to avoid the random masking strategy that removes too much or all patches from the rotated crop, we separately mask $r$ and $b$ at a certain ratio, \textit{e.g.}, 75\%. Thus, visible patches and masked patches from the background are denoted as ${b^v} = \{b_i^v|i = 1,2,...,N_b^v\}$ and ${b^m} = \{b_i^m|i=1,2,...,N_b^m\}$ respectively. These definition holds similarly for the rotated crop patches $r$.

{\bf Reconstruction.} MA3E uses MSE loss to predict the pixel values of masked patches in the background. For the rotated crop, an OT loss ${\cal L}_{OT}$ is proposed to minimize the distance in pixel space between each patch and its matched image patches. The overall loss can be written as follows:

\begin{equation}
\label{eq2}
{\cal L}_{rec} = {\cal L}_{MSE}(b^m,\hat b^m) + {\cal L}_{OT}(r,\hat r),
\end{equation}
where ${\hat b^m} \in {\mathbb R}^{N_b^m \times {p^2}C}$ means predictions for the masked patches of the background, and $\hat r \in {\mathbb R}^{{N_r} \times {p^2}C}$ denotes all predicted patches of the rotated crop.

\subsection{Reconstruction for Rotated Crop}

After the \textit{scaling center crop} operation, the scenes in all rotated crop patches are offset compared to the original image patches at the same positions. Directly calculating MSE between the masked patches and image patches not only results in biases but also overlooks the changes in angles and scenes on the visible patches. Inspired by~\cite{ge2021ota}, this paper treats the reconstruction for rotated crops as an OT problem, allowing each predicted patch to automatically match similar image patches for reconstruction.

{\bf Optimal transport.} Supposing there are $M$ suppliers and $N$ demanders, where the $i$-th supplier holds $u_i$ units of goods, and the $j$-th demander needs $v_j$ units of goods. The transportation cost for transporting one unit of goods from the $i$-th supplier to the $j$-th demander is denoted as $c_{ij}$. OT aims to find a transportation plan, denoted as $\Omega = \{\omega _{i,j}|i = 1,2,...,M,j = 1,2,...,N\}$, that minimizes the total transportation costs, ensuring that all goods are transported from suppliers to demanders:

\begin{equation}
\label{eq3}
\begin{aligned}
\min _\omega & \sum_{i=1}^M \sum_{j=1}^N c_{i j} \omega_{i j} . \\
\text {s.t.} & \sum_{i=1}^M \omega_{i j}=v_j, \sum_{j=1}^N \omega_{i j}=u_i, \\
& \sum_{i=1}^M u_i=\sum_{j=1}^N v_j, \\
& \omega_{i j} \geq 0, i=1,2, \ldots, M, j=1,2, \ldots, N .
\end{aligned}
\end{equation}

{\bf OT for reconstruction.} The context of reconstructing the rotated crop is described in Fig.~\ref{fig2}(b), considering $N_r$ original image patches and $N_r$ predicted patches of the rotated crop, each image patch is treated as a supplier, holding ${p^2}C$ units of pixel values (\textit{i.e.}, $u_i = {p^2}C, i = 1,2,...,{N_r}$), and each predicted patch as a demander, with ${p^2}C$ units of channels (\textit{i.e.}, $v_j = {p^2}C, j =1,2,...,{N_r}$) needing ${p^2}C$ units of pixel values for reconstruction. The similarity between the each unit of pixel value in the image patch and the any unit of channel in the predicted patch represents the transportation cost $c_{ij}$. This is extended to matrix-wise MSE computation for GPU acceleration. Thus, the transportation cost from the $i$-th image patch to the $j$-th predicted patch is given by:

\begin{equation}
\label{eq4}
{c_{ij}} = \left\| {{r_i} - {{\hat r}_j}} \right\|_2^2,
\end{equation}
where image patches closer in L2 distance to the predicted patch have higher similarity, tending to lower costs. A fast iterative algorithm, Sinkhorn-Knopp~\cite{cuturi2013sinkhorn}, is employed to calculate the transportation plan $\Omega$ in Eq.~\ref{eq3}. According to the solved $\Omega = \{\omega_{i,j}\}$, as shown in Fig.~\ref{fig2}(b), the OT loss automatically allocates similar multiple image patches for $j$-th predicted patch as the reconstruction targets, defined as follows:

\begin{equation}
\label{eq5}
\mathcal{L}_{OT}(r, \hat{r})=\sum_{i=1}^{N_r} \sum_{j=1}^{N_r}\left\|r_i-\hat{r}_j\right\|_2^2 \omega_{i j}
\end{equation}

The proposed ${\cal L}_{OT}$, during the pixel reconstruction and angle restoration, guides our model to perceive the angle variations of rotated crops. As the iterations progress, MA3E can effectively learn rotation-invariant visual representations. Supplementary material provides more details about ${\cal L}_{OT}$.

\section{Experiments}

Following the evaluation protocol of \cite{bao2021beit, he2022masked}, MA3E is first pre-trained on MillionAID~\cite{long2021creating}. Then, only the encoder is fine-tuned on downstream tasks, including scene classification, rotated object detection, and semantic segmentation.

\subsection{Experimental Setups}
\label{sec4.1}

\begin{table}[tb]
\centering
\rowcolors{25}{lightgreen}{lightgreen}
\begin{tabular}{lcccccccc}
\multirow{2}{*}{Methods} & \multirow{2}{*}{Backbone} & PT & \multirow{2}{*}{Ep.} & FLOPs & GPU & NU45 & AID & UCM \\
 & & Data. & & (G) & H. / Ep. & (2:8) & (5:5) & (5:5) \\
 \hline
\multicolumn{9}{l}{\textit{\textcolor{gray}{Pre-training Methods for Natural Images}}} \\
MoCo v3 $\star$~\cite{chen2021empirical}& ViT-B      & IN1k & 300  & 17.5 &  -  & 80.69 & 85.74 & 81.43 \\
MoCo v3 \dag~\cite{chen2021empirical} & ViT-B        & MA   & 300  & 17.5 & 1.9 & 92.40 & 93.99 & 97.70 \\
DINO $\star$~\cite{caron2021emerging} & ViT-B        & IN1k & 400  & 17.5 &  -  & 78.71 & 83.14 & 80.14 \\
DINO \dag~\cite{caron2021emerging}    & ViT-B        & MA   & 300  & 17.5 & 3.6 & 90.88 & 93.36 & 97.66 \\
MAE $\star$~\cite{he2022masked}       & ViT-B        & IN1k & 1600 & 17.5 &  -  & 95.20 & 97.64 & 99.05 \\
MAE \dag~\cite{he2022masked}          & ViT-B        & MA   & 300  & 17.5 & 1.2 & 95.31 & 98.16 & 99.05 \\
MAE \dag~\cite{he2022masked}          & ViT-B        & MA   & 1600 & 17.5 & 1.2 & 95.40 & 98.36 & 99.44 \\
SimMIM \dag~\cite{xie2022simmim}      & ViT-B        & MA   & 400  & 17.5 & 2.7 & 95.54 & 98.19 & 99.06 \\
LoMaR \dag~\cite{chen2022efficient}   & ViT-B        & MA   & 300  & 17.5 & 1.9 & 95.47 & 98.11 & 98.91 \\
MixMAE \dag~\cite{liu2023mixmae}      & Swin-B/W14   & MA   & 300  & 16.3 & 1.8 & 95.45 & 98.22 & 99.04 \\
\hline
\multicolumn{9}{l}{\textit{\textcolor{gray}{Pre-training Methods for RS Images}}} \\
SeCo $\star$~\cite{manas2021seasonal} & RN-50        & S-2  & 200  & 4.1  &  -  & 92.91	& 95.99	& 97.81 \\
CACo $\star$~\cite{mall2023change}    & RN-50        & S-2  & 200  & 4.1  &  -  & 91.94	& 95.05	& 97.05 \\
RingMo~\cite{sun2022ringmo}           & Swin-B & $\sim$2M   & 200  & 15.6 &  -  & 95.67 & 98.34 &   -   \\
CMID $\star$~\cite{muhtar2023cmid}    & Swin-B       & MA   & 200  & 15.6 &  -  & 95.16 & 96.98 & 98.21 \\
GFM $\star$~\cite{mendieta2023towards}& Swin-B       &G.I.  & 100  & 15.6 &  -  & 96.06 & 97.09 & 99.14 \\
MAE~\cite{wang2022advancing}          & ViT-B+RVSA   & MA   & 1600 & 33.6 &  -  & 95.49 & 98.33 & 99.70 \\
MAE~\cite{wang2022advancing}          & ViTAE-B+RVSA & MA   & 1600 & 26.3 &  -  & 95.69 & 98.48 & 99.56 \\
SatMAE $\star$~\cite{cong2022satmae}  & ViT-B        & f-S  & 200  & 17.5 &  -  & 76.04 & 83.84 & 81.05 \\
SatMAE $\star$~\cite{cong2022satmae}  & ViT-L        & f-R  & 800  & 61.3 &  -  & 93.78 & 98.70 & 97.14 \\
ScaleMAE $\star$~\cite{reed2023scale} & ViT-L        & f-R  & 800  & 61.3 &  -  & 88.54 & 97.42 & 93.28 \\
\hline
MA3E                                  & ViT-B        & MA   & 300  & 17.5 & 1.4 & 95.77 & 98.44 & 99.05 \\
MA3E                                  & ViT-B        & MA   & 1600 & 17.5 & 1.4 & \textbf{96.23} & \textbf{99.04} & \textbf{99.81} \\
\end{tabular}
\caption{Comparison of fine-tuning results on three scene classification datasets. The FLOPs are evaluated for each backbone in one GPU. \dag \ indicates our reproduction with the official code. Due to the limited GPU memory, we adjust the \textit{batchsize} per GPU and accumulation steps to maintain their default global \textit{batchsize}. $\star$ denotes direct fine-tuning the released pre-trained weights. Dataset abbreviations: IN1k for ImageNet-1k, MA for MillionAID, S-2 for Sentinel-2~\cite{drusch2012sentinel}, $\sim$2M for about two million images, G.I. for GeoPile~\cite{mendieta2023towards} and ImageNet-22k, f-S for fMoW-Sentinel~\cite{christie2018functional}, f-R for fMoW-RGB~\cite{christie2018functional}.}
\label{tab_cls_ft}
\end{table}

\begin{table}[t]
\centering
\rowcolors{23}{lightgreen}{lightgreen}
\begin{tabular}{lcccccc}
\multirow{2}{*}{Methods} & \multirow{2}{*}{Backbone} & Pre-training & \multirow{2}{*}{Epoch} & NU45 & AID & UCM \\
 & & Dataset & & (2:8) & (5:5) & (5:5) \\
\hline
\multicolumn{7}{l}{\textit{\textcolor{gray}{Pre-training Methods for Natural Images}}} \\
MoCo v3 $\star$~\cite{chen2021empirical}& ViT-B      & ImageNet-1k   & 300  & 48.80 & 68.51 & 25.40 \\
MoCo v3 \dag~\cite{chen2021empirical} & ViT-B        & MillionAID    & 300  & 61.45 & 78.72 & 38.34 \\
DINO $\star$~\cite{caron2021emerging} & ViT-B        & ImageNet-1k   & 400  & 51.42 & 72.10 & 31.14 \\
DINO \dag~\cite{caron2021emerging}    & ViT-B        & MillionAID    & 300  & 63.67 & 78.51 & 40.04 \\
MAE $\star$~\cite{he2022masked}       & ViT-B        & ImageNet-1k   & 1600 & 66.09 & 74.60 & 49.81 \\
MAE \dag~\cite{he2022masked}          & ViT-B        & MillionAID    & 300  & 73.94 & 83.12 & 51.52 \\
MAE \dag~\cite{he2022masked}          & ViT-B        & MillionAID    & 1600 & 75.98 & 84.21 & 52.75 \\
SimMIM \dag~\cite{xie2022simmim}      & ViT-B        & MillionAID    & 400  & 74.86 & 83.19 & 51.48 \\
LoMaR \dag~\cite{chen2022efficient}   & ViT-B        & MillionAID    & 300  & 74.30 & 82.26 & 51.89 \\
MixMAE \dag~\cite{liu2023mixmae}      & Swin-B/W14   & MillionAID    & 300  & 73.95 & 81.53 & 50.63 \\
\hline
\multicolumn{7}{l}{\textit{\textcolor{gray}{Pre-training Methods for RS Images}}} \\
SeCo $\star$~\cite{manas2021seasonal} & RN-50        & Sentinel-2    & 200  & 65.02	& 78.26	& 47.45 \\
CACo $\star$~\cite{mall2023change}    & RN-50        & Sentinel-2    & 200  & 63.24	& 77.81	& 40.53 \\
CMID $\star$~\cite{muhtar2023cmid}    & Swin-B       & MillionAID    & 200  & 65.63 & 79.05 & 47.43 \\
GFM $\star$~\cite{mendieta2023towards}& Swin-B       & GeoPile+ImageNet-22k & 100  & 76.09 & 80.58 & 49.73 \\
MAE $\star$~\cite{wang2022advancing}  & ViT-B+RVSA   & MillionAID    & 1600 & 75.72 & 84.06 & 50.86 \\
SatMAE $\star$~\cite{cong2022satmae}  & ViT-B        & fMoW-Sentinel & 200  & 20.60 & 33.72 & 19.14 \\
SatMAE $\star$~\cite{cong2022satmae}  & ViT-L        & fMoW-RGB      & 800  & 37.15 & 55.10 & 34.28 \\
ScaleMAE $\star$~\cite{reed2023scale} & ViT-L        & fMoW-RGB      & 800  & 33.03 & 48.46 & 28.19 \\
\hline
MA3E                                  & ViT-B        & MillionAID    & 300  & 74.61 & 84.21 & 52.24 \\
MA3E                                  & ViT-B        & MillionAID    & 1600 & \textbf{76.41} & \textbf{85.86} & \textbf{55.69} \\
\end{tabular}
\caption{Comparison of linear probing results on three scene classification datasets. \dag \ indicates our reproduction with the official code. $\star$ denotes direct linear probing the released pre-trained weights.}
\label{tab_cls_lin}
\end{table}

Unless otherwise stated, all experiments are implemented using PyTorch and conducted on a machine equipped with eight 24GB RTX 3090 GPUs. More experimental setups and datasets are detailed in the supplementary material.

{\bf Pre-training details.} The testing set of MillionAID~\cite{long2021creating}, consisting of 990,848 RS images, is used for pre-training. Each image are resized to $224 \times 224$ pixels. The patch size $p$ is 16. For each rotated crop, the side length $a$ is set to 96, and the rotation range is $[-45^\circ, +45^\circ]$. We randomly mask the rotated crop and background patches with a ratio of 75\% respectively. MA3E employs a plain ViT-B~\cite{dosovitskiy2020image} as the encoder and 8 ViT blocks with 512-D as the decoder. Except for the \textit{batchsize} of 1024, all other pre-training configurations follow \cite{he2022masked}.

{\bf Scene classification.} All fine-tuning and linear probing experiments are conducted on NWPU-RESISC45~\cite{cheng2017remote} (NU45), AID~\cite{xia2017aid}, and UC Merced~\cite{yang2010bag} (UCM) datasets. For NU45, 20\% of images from each class are randomly sampled as the training set, and the remaining 80\% are used for testing. For AID and UCM, these two ratios are both 50\%. Fine-tuning is performed with a \textit{batchsize} of 512 for 200 \textit{epochs}, and linear probing is trained with a \textit{batchsize} of 2048 for 100 \textit{epochs}. We follow the other default fine-tuning and linear probing settings outlined in \cite{he2022masked} and report the Top-1 accuracy on each testing set. 

{\bf Rotated object detection.} Experiments for detection are conducted on DOTA1.0~\cite{xia2018dota} and DIOR-R~\cite{cheng2022anchor} using the Oriented R-CNN~\cite{xie2021oriented} detector. MA3E pre-trained models serves as the backbone of the detector and undergoes end-to-end fine-tuning. The detector uses the \textit{batchsize} of 2 for DOTA1.0 and 4 for DIOR-R. We train for 12 \textit{epochs}, with other hyper-parameters following default settings of the detector. Mean Average Precision (mAP) on each testing set is reported, where results on DOTA1.0 are obtained from the official evaluation server. Due to the limited GPU memory, above experiments are deployed on two GPUs and implemented by the \href{https://github.com/jbwang1997/OBBDetection}{OBBDetection} and the ViTDet~\cite{li2022exploring} codebases.

{\bf Semantic segmentation.} Similarly, segmentation experiments are conducted using the UperNet~\cite{xiao2018unified} framework for end-to-end supervised fine-tuning on iSAID~\cite{waqas2019isaid} and \href{https://www.isprs.org/education/benchmarks/UrbanSemLab/2d-sem-label-potsdam.aspx}{Potsdam}. The UperNet is trained for $160k$ iterations with a \textit{batchsize} of 4, while other hyper-parameters remain at the default settings. Mean Intersection over Union (mIoU) on the iSAID validation set and mean F1 score (mF1) on the Potsdam testing set are reported. These experiments are implemented using the mmsegmentation~\cite{contributors2020mmsegmentation} library and also run on two GPUs.

\begin{table}[t]
\centering
\rowcolors{25}{lightgreen}{lightgreen}
\begin{tabular}{lcccccp{0.5pt}cc}
\multirow{2}{*}{Methods} & \multirow{2}{*}{Backbone} & PT & \multirow{2}{*}{Ep.} & DOTA1.0 & DIOR-R && iSAID & Potsdam \\
\cline{5-6} \cline{8-9}
 & & Data. &  & \multicolumn{2}{c}{mAP} && mIoU & mF1 \\
\hline
\multicolumn{9}{l}{\textit{\textcolor{gray}{Pre-training Methods for Natural Images}}} \\
MoCo v3 $\star$~\cite{chen2021empirical}& ViT-B      & IN1k  & 300  & 59.35 & 44.22 && 40.18 & 83.59 \\
MoCo v3 \dag~\cite{chen2021empirical} & ViT-B        & MA    & 300  & 71.46 & 59.41	&& 58.72 & 90.13 \\
DINO $\star$~\cite{caron2021emerging} & ViT-B        & IN1k  & 400  & 73.53	& 62.67	&& 50.40 & 86.29 \\
DINO \dag~\cite{caron2021emerging}    & ViT-B        & MA    & 300  & 74.91	& 64.87	&& 54.61 & 88.56 \\
MAE $\star$~\cite{he2022masked}       & ViT-B        & IN1k  & 1600 & 76.04 & 64.84 && 61.08 & 90.14 \\
MAE \dag~\cite{he2022masked}          & ViT-B        & MA    & 300  & 75.85 & 64.54 && 60.96 & 90.08 \\
MAE \dag~\cite{he2022masked}          & ViT-B        & MA    & 1600 & 77.53 & 67.72 && 61.38 & 90.49 \\
SimMIM \dag~\cite{xie2022simmim}      & ViT-B        & MA    & 400  & 76.17 & 65.24 && 60.92 & 90.20 \\
LoMaR \dag~\cite{chen2022efficient}   & ViT-B        & MA    & 300  & 75.76 & 64.55 && 60.86 & 90.21 \\
MixMAE \dag~\cite{liu2023mixmae}      & Swin-B/W14   & MA    & 300  & 75.87 & 64.67 && 60.64 & 90.13 \\
\hline
\multicolumn{9}{l}{\textit{\textcolor{gray}{Pre-training Methods for RS Images}}} \\
SeCo $\star$~\cite{manas2021seasonal} & RN-50        & S-2   & 200  & 69.95	& 62.74	&& 57.45 & 89.83 \\
CACo $\star$~\cite{mall2023change}    & RN-50        & S-2   & 200  & 75.35	& 65.10 && 61.32 & 90.35 \\
CMID~\cite{muhtar2023cmid}            & Swin-B       & MA    & 200  & 77.36 &   -   &&   -   &   -   \\
CMID $\star$~\cite{muhtar2023cmid}    & Swin-B       & MA    & 200  & 77.29 & 66.13 && 62.42 & 90.71 \\
GFM $\star$~\cite{mendieta2023towards}& Swin-B       & G.I.  & 100  & 77.81	& 67.67	&& 62.54 & 90.62 \\
MAE~\cite{wang2022advancing}          & ViT-B+RVSA   & MA    & 1600 & 78.75 & 70.67 && 63.76 & 90.60 \\
MAE~\cite{wang2022advancing}          & ViTAE-B+RVSA & MA    & 1600 & 78.96 & 70.95 && 63.48 & 91.22 \\
SatMAE $\star$~\cite{cong2022satmae}  & ViT-B        & f-S   & 200  & 68.54 & 48.55 && 53.55 & 86.43 \\
\hline
MA3E                                  & ViT-B        & MA    & 300  & 77.93 & 68.41 && 62.74 & 90.67 \\
MA3E                                  & ViT-B        & MA    & 1600 & \textbf{79.47} & \textbf{71.82} && \textbf{64.06} & \textbf{91.50}
\end{tabular}
\caption{Comparison of rotated object detection and semantic segmentation results on different datasets. \dag \ indicates our reproduction with the available code using MillionAID. $\star$ denotes end-to-end fine-tuning by directly loading the released pre-trained weights. We only fine-tune the base-level models due to limited GPU memory.}
\label{tab_detseg}
\end{table}

\subsection{Main Results}

MA3E is compared with eight state-of-the-art pre-training methods for RS images, including seasonal-contrasted SeCo~\cite{manas2021seasonal}, change-aware contrasted CACo~\cite{mall2023change}, GFM~\cite{mendieta2023towards} with continual pre-training, RingMo~\cite{sun2022ringmo} adopting incomplete masking, CMID~\cite{muhtar2023cmid} combining contrastive learning and MIM, \cite{wang2022advancing} using rotated varied-size window attention (RVSA) to replace the original global attention during downstream fine-tuning, sptaio-temporal encoded SatMAE~\cite{cong2022satmae}, and scale-aware ScaleMAE~\cite{reed2023scale}. However, these methods adopt different datasets and fine-tuning settings for downstream tasks. For fairness, we normalize the experimental setups and further compared with six pre-training methods for natural images, including the popular MoCo v3~\cite{chen2021empirical}, DINO~\cite{caron2021emerging}, MAE~\cite{he2022masked}, SimMIM~\cite{xie2022simmim}, region-reconstructed LoMaR~\cite{chen2022efficient}, and input-mixed MixMAE~\cite{liu2023mixmae}.

{\bf Scene classification.} The fine-tuning and linear probing results on three datasets are shown in Table~\ref{tab_cls_ft} and Table~\ref{tab_cls_lin}, repectively. MA3E pre-trained for 300 \textit{epochs} achieves competitive results. Although the fine-tuning accuracy on UCM is lower than ViT+RVSA~\cite{wang2022advancing}, MA3E requires only 52\% of the latter's FLOPs. The fine-tuning and linear probing results on three datasets continually improve as training progresses. MA3E pre-trained for 1600 \textit{epochs} comprehensively leads, demonstrating that MA3E effectively learns the discriminative rotation-invariant representations of RS objects. In addition, the training time per \textit{epoch} on a single GPU is calculated. Compared to MAE~\cite{he2022masked}, we achieve a significant improvement in accuracy with only about 0.2 hours of extra training time.

\begin{table}[htbp]
\begin{minipage}{0.58\textwidth}
\centering
\begin{tabular}{cccccccc}
MAE        & SCC        & AE         & Mask.      & OT         & ft    & det   & seg \\
\hline
\checkmark &            &            &            &            & 95.31 & 75.85 & 60.96 \\
\checkmark & \checkmark &            &            &            & 95.43 & 76.12 & 61.24 \\
\checkmark & \checkmark & \checkmark &            &            & 95.47 & 76.41 & 61.86 \\
\checkmark & \checkmark &            & \checkmark &            & 95.36 & 76.46 & 61.88 \\
\checkmark & \checkmark &            &            & \checkmark & 95.06 & 77.23 & 62.17 \\
\checkmark & \checkmark & \checkmark & \checkmark &            & 95.53 & 76.70 & 61.93 \\
\checkmark & \checkmark & \checkmark & \checkmark & \checkmark & \cellcolor{lightgreen}\textbf{95.77} & \cellcolor{lightgreen}\textbf{77.93} & \cellcolor{lightgreen}\textbf{62.74}
\end{tabular}
\caption{The ablation results of MA3E using different components. SCC: \textit{scaling center crop} operation, AE: angle embedding, Mask.: random masking the rotated crop and background respectively.}
\label{tab_ablation}
\end{minipage}
\hfill
\begin{minipage}{0.4\textwidth}
\centering
\begin{tabular}{ccccc}
$a$ & Num. & ft & det & seg \\
\hline
32 & 1 & 95.21 & 76.13 & 61.89 \\
32 & 4 & 94.16 & 75.59 & 62.14 \\
64 & 1 & 95.36 & 77.53 & 62.51 \\
64 & 2 & 94.90 & 76.21 & 62.43 \\
80 & 1 & 95.40 & 77.38 & \textbf{62.83} \\
96 & 1 & \cellcolor{lightgreen}\textbf{95.77} & \cellcolor{lightgreen}\textbf{77.93} & \cellcolor{lightgreen}62.74 \\
128 & 1 & 95.74 & 76.92 & 62.01
\end{tabular}
\caption{The side length $a$ of the rotated crop and the numbers of non-overlapping rotated crops in a composite image.}
\label{tab_side_l_crop_nums}
\end{minipage}
\end{table}

{\bf Rotated object detection.} Table~\ref{tab_detseg} presents the fine-tuning results on DOTA1.0 and DIOR-R for different methods, where MA3E, using a simple backbone, obtains superior detection performance. The version pre-trained for 300 \textit{epochs} outperforms other methods with a similar number of \textit{epochs}. MA3E pre-trained for 1600 \textit{epochs} surpasses all methods. Compared to \cite{wang2022advancing} with ViTAE+RVSA, the mAP on DOTA1.0 and DIOR-R has increased by 0.51 and 0.87, respectively. This significant improvement in detection performance demonstrates the effectiveness of MA3E in angle perception during pre-training.

{\bf Semantic segmentation.} The fine-tunning results on iSAID and Potsdam are also reported in Table~\ref{tab_detseg}. With fewer pre-training \textit{epochs}, MA3E demonstrates competitive performance, achieving mF1 only 0.04 lower than CMID~\cite{muhtar2023cmid} on Potsdam. When pre-trained for 1600 \textit{epochs}, MA3E again achieves the best results, outperforming the second-best~\cite{wang2022advancing} by 0.3 mIoU and 0.28 mF1 on iSAID and Potsdam, respectively. It is indicated that the significance of the rotation-invariant representations learned by our model in semantic segmentation.

\subsection{Ablation Study}

\begin{table}[t]
\begin{minipage}{0.49\textwidth}
\centering
\begin{tabular}{lccc}
Position & ft & det & seg \\
\hline
fixed & 95.48 & 76.44 & 62.10 \\
random  & \cellcolor{lightgreen}95.77 & \cellcolor{lightgreen}77.93 & \cellcolor{lightgreen}\textbf{62.74} \\
selective search~\cite{uijlings2013selective} & \textbf{95.96} & \textbf{78.08} & 62.68
\end{tabular}
\caption{How to determine the position of the rotated crop.}
\label{tab_crop_pos}
\end{minipage}
\hfill
\begin{minipage}{0.49\textwidth}
\centering
\begin{tabular}{lccc}
Operation & ft & det & seg \\
\hline
random rotation  & 93.82 & 76.14 & 61.23 \\	
scaling center crop & \cellcolor{lightgreen}\textbf{95.77} & \cellcolor{lightgreen}\textbf{77.93} & \cellcolor{lightgreen}\textbf{62.74} \\
 & & & 
\end{tabular}
\caption{How to create the rotated crop.~\\~\\}
\label{tab_crop_types}
\end{minipage}
\end{table}

\begin{table}[t]
\begin{minipage}{0.49\textwidth}
\centering
\begin{tabular}{lccc}
Range & ft & det & seg \\
\hline
$[-30^\circ, +30^\circ]$ & \textbf{95.78} & 77.68 & 62.49 \\
$[-45^\circ, +45^\circ]$ & \cellcolor{lightgreen}95.77 & \cellcolor{lightgreen}\textbf{77.93} & \cellcolor{lightgreen}\textbf{62.74} \\
$[-60^\circ, +60^\circ]$ & 95.32 & 77.22 & 62.55 \\
$[-90^\circ, +90^\circ]$ & 94.89 & 76.45 & 61.90
\end{tabular}
\caption{Rotation range.}
\label{tab_max_r}
\end{minipage}
\hfill
\begin{minipage}{0.49\textwidth}
\centering
\begin{tabular}{lccc}
Strategy & ft & det & seg \\
\hline
random masking & \cellcolor{lightgreen}\textbf{95.77} & \cellcolor{lightgreen}\textbf{77.93} & \cellcolor{lightgreen}\textbf{62.74} \\
block-wise masking & 94.98 & 77.69 & 62.39 \\
uniform sampling~\cite{li2022uniform} & 95.33 & 77.57 & 62.58 \\
 & & & 
\end{tabular}
\caption{Masking strategy.}
\label{tab_mask_strategy}
\end{minipage}
\end{table}

In this section, a series of ablation studies are conducted to analyze how each key design is available and demonstrate the effectiveness of each component in MA3E. By default, MA3E is pre-trained for 300 \textit{epochs}. We report the Top-1 accuracy on NU45~\cite{cheng2017remote} after fine-tuning, mAP on DOTA1.0~\cite{xia2018dota}, and mIoU on iSAID~\cite{waqas2019isaid}. The results of MA3E with default settings are marked in \colorbox{lightgreen}{green}. Our supplementary material provides additional ablation results.

{\bf Each component.} The ablation results on different components are shown in Table~\ref{tab_ablation}. Each key design of MA3E improves the performance of the baseline MAE across the three RS tasks, and the combination of all components in MA3E yields the best results. This demonstrates the effectiveness of our proposal.

{\bf Side length $a$ and crop numbers.} We study the effects of different side lengths $a$ for the rotated crop and the number rotated crops in each image, as shown in Table~\ref{tab_side_l_crop_nums}. As $a$ increases, MA3E's performance gradually improves until $a=96$. However, for detection and segmentation tasks, there is a sudden decrease in the performance when $a=128$. Furthermore, an increase in the number of rotated crops also negatively impacts performance. Excessive large $a$ and a more number of crops make angle restoration challenging, and this decline is more obvious in detection where angles are crucial.

{\bf Position of rotated crops.} We ablate three schemes for selecting the position of the rotated crop: fixed positions at the center of the images, random positions, and positions determined by the selective search~\cite{uijlings2013selective}. The corresponding results are shown in Table~\ref{tab_crop_pos}. The selective search algorithm generates candidate bounding boxes for potential objects in an unsupervised manner. Although determining the position of the rotated crop through this method results in an improvement of 0.19 in classification accuracy and 0.15 mAP in detection compared to randomly choosing positions, the limited performance gain comes at the cost of at least 10\% extra training time per \textit{epoch}. Therefore, it is not employed in the default settings.

{\bf Method to create rotated crops.} We compare MA3E with a model using the simple \textit{random rotation} operation to generate rotated crops with $a=96$. The results in Table~\ref{tab_crop_types} demonstrate that rotated crops constructed using the proposed \textit{scaling center crop} operation significantly enhance the performance in all tasks. Note that our method may still incur minor losses in edge scenes. But this corresponds to deliberately increasing the difficulty of reconstructing the complete original image, thereby improving the quality of learned rotation-invariant representations.

{\bf Rotation ranges.} Table~\ref{tab_max_r} shows the results for different rotation ranges of rotated crops. An reasonable range maximizes the ability of ${\cal L}_{OT}$ to restore the angle variations introduced on rotated crops, promoting MA3E to learn rotation-invariant visual representations and achieving good performance.

{\bf Masking strategy.} Table~\ref{tab_mask_strategy} report the effect of different masking strategies on MA3E. The random masking strategy~\cite{he2022masked} performs better. Block-wise masking~\cite{bao2021beit} increases the reconstruction difficulty. Meanwhile, uniform sampling~\cite{li2022uniform} with masking in adjacent four patches makes the reconstruction of rotated crops easier but leads to lower quality representations learned by the model.

\subsection{Visualization}

\begin{figure}[t]
\centering
\includegraphics[width=4.8in]{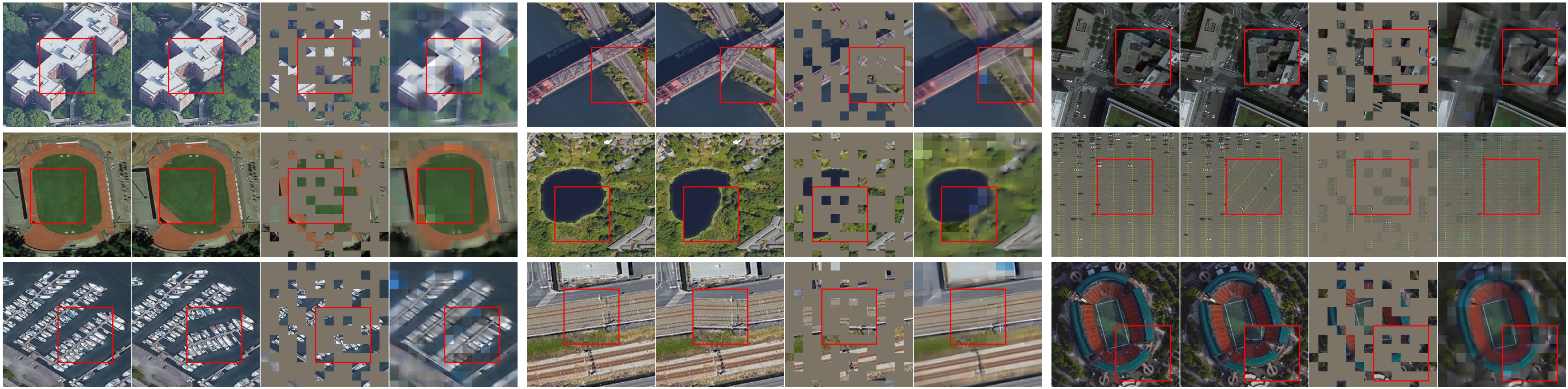}
\caption{Example results on MillionAID training images. For each set, we display the original image (left one), the composite image containing the rotated crop (left two), the masked image (right two), and the MA3E reconstructed image (right one). To aid observation, the rotated crop is highlighted with a red box. Following MAE~\cite{he2022masked}, we show the model's output on visible patches to comprehensively demonstrate the reconstruction quality of MA3E.}
\label{fig4}
\end{figure}

Fig.~\ref{fig4} quantitatively visualizes the reconstruction performance of MA3E on RS images. Example images are randomly sampled from the training set of MillionAID~\cite{long2021creating}, which contains $10k$ images. We resize each image to $224 \times 224$ (196 patches, $p=16$) and set the rotated crop of $96 \times 96$ (36 patches) for reconstruction. The rotated crop and background are masked at a 75\% ratio, corresponding to 9 and 40 visible patches, respectively. It can be seen that MA3E effectively restores the preset angle variations of the rotated crops during reconstructing original pixels. In addition, some uneven color patches may appear in the reconstructed rotated crop, such as the water surface and playground in Fig.~\ref{fig4}. This phenomenon is attributed to the proposed OT loss, which calculates the mean square error between each predicted patch and multiple target patches. The supplementary material shows more visualizations.

\section{Conclusion and Discussion}

This paper proposes Masked Angle-Aware Autoencoder (MA3E) for self-supervised representation learning on RS images. The \textit{scaling center crop} operation is designed to construct the rotated crop within each original image, introducing the explicit angle variation. MA3E takes the created composite image as input, with the goal of simultaneously achieving original pixel reconstruction and angle restoration. The reconstruction for the rotated crop is treated as an optimal transport problem, and we propose an OT loss to automatically assign similar original image patches for each rotated crop patch. Finally, MA3E can effectively perceive angles and learn rotation-invariant representations, achieving competitive performance in various downstream tasks. We hope MA3E can contribute to the advancement of foundational models in RS research. 

{\bf Limitation.} Despite MA3E demonstrates the potential of angle awareness, in many RS scenarios, the angles of only man-made objects require more emphasis. For extensive land cover, the model may not benefit significantly from angle information. In future work, we will consider the scale that exists in any RS scene and further explore the combination of angles for man-made objects and scale for land cover.

\section*{Acknowledgements}
This work is supported in part by the National Natural Science Foundation of China under Grant 62171347, 62101405, 62371373, 62271377; 111 Project; the Postdoctoral Fellowship Program of China Postdoctoral Science Foundation under Grant GZC20232036, GZC20232032; the Shaanxi Province postdoctoral research project under Grant 2023BSHEDZZ168, 2023BSHYDZZ96; the Fundamental Research Funds for the Central Universities.

%
%
\bibliographystyle{splncs04}


\newpage
\title{Masked Angle-Aware Autoencoder for Remote Sensing Images} 
\titlerunning{Masked Angle-Aware Autoencoder}

\author{Zhihao Li\orcidlink{0000-0001-7119-3215} \and
Biao Hou$^\dag$ \orcidlink{0000-0002-1996-186X} \and
Siteng Ma\orcidlink{0000-0001-9678-0213} \and
Zitong Wu\orcidlink{0000-0002-0449-9465} \and 
Xianpeng Guo\orcidlink{0000-0003-3733-2570} \and
Bo Ren\orcidlink{0000-0002-0481-5069} \and
Licheng Jiao\orcidlink{0000-0003-3354-9617}}

\authorrunning{Z. Li et al.}

\institute{School of Artificial Intelligence, Xidian University \\
\email{avcodec@163.com} \\
$^\dag$ Corresponding Author
}

\appendix

\section{Full Implementation Details}

\subsection{Experimental Setups}

{\bf Pre-training.} The pre-training settings of MA3E is in Table.~\ref{tab_pretrain_lp}. All ViT blocks are initialized by xavier\_uniform~\cite{glorot2010understanding}. We use the \textit{batchisze} of 1024 and the linear \textit{lr} scaling rule~\cite{goyal2017accurate}: $lr = base \ lr \times batchsize / 256$. Simple data augmentations such as random cropping and horizontal flipping are also applied before creating rotated crops.

\begin{table}[htpb]
\centering
\begin{tabular}{l|c|c}
configs   & pre-training & linear probing \\
\hline
optimizer & AdamW~\cite{loshchilov2017decoupled} & LARS~\cite{you2017large} \\
base learning rate & 1.5e-4 & 0.1 \\
weight decay & 0.05 & 0 \\
optimizer momentum & $\beta_1, \beta_2 = 0.9, 0.95$~\cite{chen2020generative} & 0.9 \\
learning rate schedule & cosine decay~\cite{loshchilov2016sgdr} & cosine decay \\
warmup epochs~\cite{goyal2017accurate} & 40 & 10 \\
augmentation & ScalingCenterCrop & RandomResizedCrop
\end{tabular}
\caption{Pre-training and linearing probing settings of MA3E.}
\label{tab_pretrain_lp}
\end{table}

{\bf Fine-tuning and linear probing.} We fine-tune for 200 \textit{epochs} with a \textit{batchsize} of 512 and linear probe for 100 \textit{epochs} with a \textit{batchsize} of 2048. Other default linear probing and fine-tuning settings are respectively shown in Table.~\ref{tab_pretrain_lp} and Table.~\ref{tab_ft}, which also follow that of MAE~\cite{he2022masked}.

\begin{table}[htpb]
\centering
\begin{tabular}{l|c}
configs   & value \\
\hline
optimizer & AdamW \\
base learning rate & 1e-3 \\
weight decay & 0.05 \\
optimizer momentum & $\beta_1, \beta_2 = 0.9, 0.999$ \\
layer-wise lr decay~\cite{bao2021beit, clark2020electra} & 0.75 \\
learning rate schedule & cosine decay \\
warmup epochs & 5 \\
augmentation & RandAug (9, 0.5)~\cite{cubuk2020randaugment} \\
label smoothing~\cite{szegedy2016rethinking} & 0.1 \\
mixup~\cite{zhang2017mixup} & 0.8 \\
cutmix~\cite{yun2019cutmix} & 1.0 \\
drop path~\cite{huang2016deep} & 0.1
\end{tabular}
\caption{End-to-end fine-tuning settings.}
\label{tab_ft}
\end{table}

{\bf Fine-tuning on DOTA1.0 and DIOR-R.} The fine-tuning details for rotated object detection is shown in Table.~\ref{tab_ft_det}. We adopt a multi-step scheduler to adjust the learning rate, which is reduced by $10 \times$ at the $8\text{-}th$ and $11\text{-}th$ \textit{epoch}.

{\bf Fine-tuning on iSAID and Potsdam.} The implementation details for fine-tuning on semantic segmentation datatsets is in Table.~\ref{tab_ft_seg}. The learning rate schedule adopts the polynomial decay policy with a \textit{power} of 1.0 and \textit{min\_lr} of 0, following \cite{wang2022advancing}.

\begin{table}[t]
\begin{minipage}{0.49\textwidth}
\centering
\begin{tabular}{l|c}
configs   & value \\
\hline
optimizer & AdamW \\
base learning rate & 1e-4 \\
weight decay & 0.05 \\
optimizer momentum & $\beta_1, \beta_2 = 0.9, 0.999$ \\
learning rate schedule & multi-step scheduler \\
drop path & 0.15 \\
\multicolumn{2}{l}{\ } \\
\multicolumn{2}{l}{\ } 
\end{tabular}
\caption{End-to-end fine-tuning settings for rotated object detection.}
\label{tab_ft_det}
\end{minipage}
\hfill
\begin{minipage}{0.49\textwidth}
\centering
\begin{tabular}{l|c}
configs   & value \\
\hline
optimizer & AdamW \\
base learning rate & 6e-5 \\
weight decay & 0.05 \\
optimizer momentum & $\beta_1, \beta_2 = 0.9, 0.999$ \\
layer-wise lr decay & 0.9 \\
learning rate schedule & polynomial scheduler \\
warmup iters & 1500 \\
drop path & 0.1
\end{tabular}
\caption{End-to-end fine-tuning settings for semantic segmentation.}
\label{tab_ft_seg}
\end{minipage}
\end{table}

\subsection{Dataset Preparations}

{\bf MillionAID~\cite{long2021creating}} is a large-scale RS scene dataset containing 1,000,848 RGB images collected from Google Earth. The training set consists of 10,000 images categorized into 51 classes, while the testing set includes the remaining 990,848 images without labels. These images are captured by various sensors and therefore have different resolutions, ranging from $110 \times 110$ to $31,672 \times 31,672$ pixels.

{\bf NWPU-RESISC45~\cite{cheng2017remote}} is a common RS image benchmark collected by Northwestern Polytechnical University from Google Earth. It contains 31,500 images in RGB color space, which are equally divided into 45 classes, each with 700 images of $256 \times 256$ pixels.

{\bf AID~\cite{xia2017aid}} has images from different countries on Google Earth. These images are extracted at different times and seasons under different imaging conditions. The dataset contains 10,000 images with $600 \times 600$ pixels in 30 classes.

{\bf UC Merced~\cite{yang2010bag}} contains 21 land-use classes, and each category has 100 images with the size of $256 \times 256$ pixels. There are a total of 2,100 RGB images from the United States Geological Survey (USGS) National Map.

{\bf DOTA1.0~\cite{xia2018dota}} is a large-scale rotated object detection dataset. It contains 2,806 images ranging from $800 \times 800$ to $4,000 \times 4,000$ pixels and has 188,282 instances with rotated bounding box annotations belonging to 15 object classes. The training, validation, and testing sets have 1,411, 458, and 937 images, respectively. Each image is cropped to $1,024 \times 1,024$ patches with a stride of 824 and both training and validation sets are used for training. Note that the testing set does not have published labels, evaluation metrics are obtained by submitting predictions on the testing set to the official evaluation server.

{\bf DIOR-R~\cite{cheng2022anchor}} is a rotated object detection dataset consisting of 20 classes. It comprises 23,463 images with a total of 192,518 instances. The training set and the testing set consist of 11,725 images with 68,073 instances and 11,738 images with 124,445 instances, respectively. All images are cropped to $800 \times 800$ pixels, with resolutions ranging from 0.5 to 30 m.

{\bf iSAID~\cite{waqas2019isaid}} is a large-scale instance segmentation dataset. Note that this dataset and DOTA1.0 share the same scenes, the difference is that iSAID is labeled with a semantic mask containing one background and 15 foregrounds class. It also consists of 2,806 high-resolution images with pixel dimension ranging from $800 \times 800$ to $4,000 \times 13,000$. We crop all images to $896 \times 896$ patches with a stride of 512 and use only the validation set for evaluation since the testing set is unavailable.

{\bf \href{https://www.isprs.org/education/benchmarks/UrbanSemLab/2d-sem-label-potsdam.aspx}{Potsdam}} is released by ISPRS Commission WG II/4. It contains 38 images with an average size of $6,000 \times 6,000$ pixels. These images cover a 3.42 km$^2$ area of Potsdam city and include six scenes, \textit{i.e.}, Impervious surface, Building, Low vegetation, Tree, Car, and Clutter. The training and testing sets have 24 and 14 images, respectively. Each image is cropped to $512 \times 512$ patches with a stride of 384. We exclude the clutter class from the dataset when calculating evaluation metrics.

\begin{figure}[!t]
\centering
\includegraphics[width=4.8in]{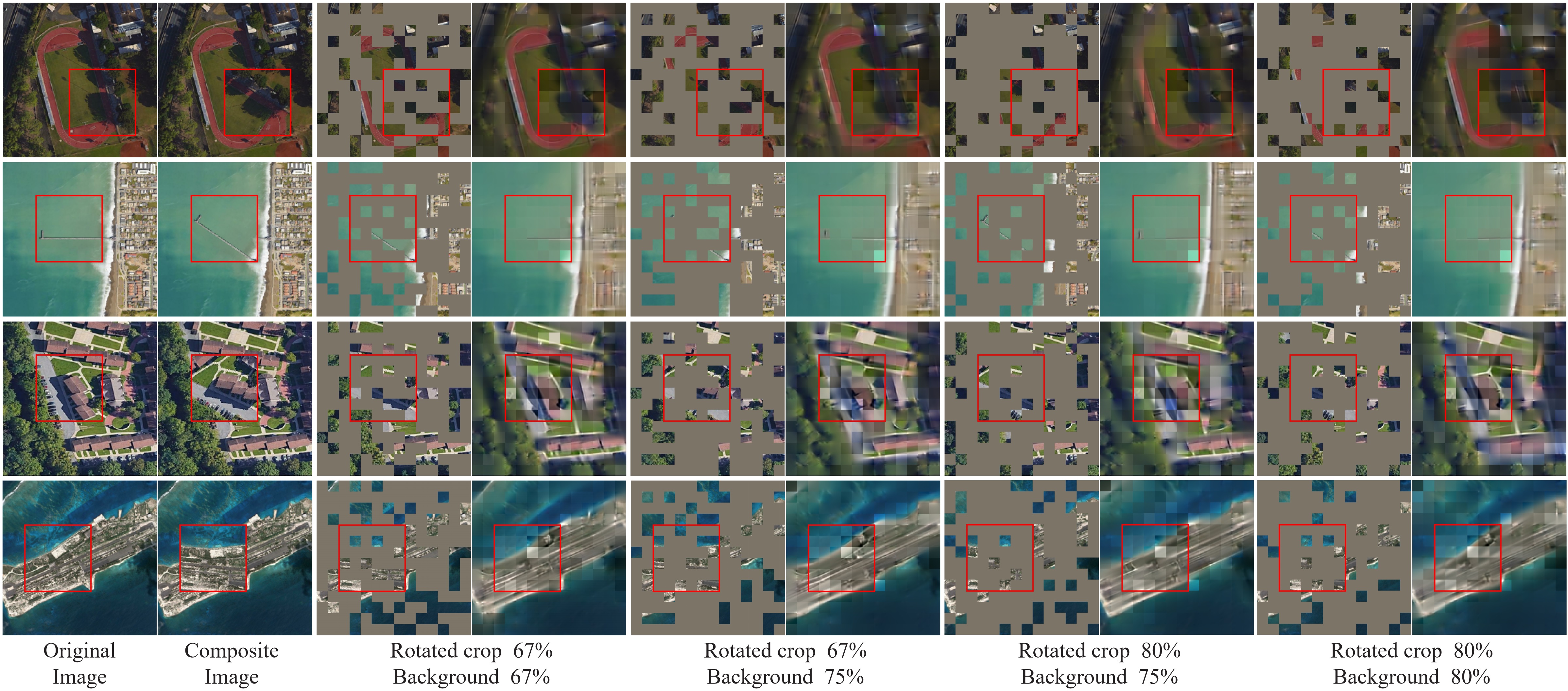}
\caption{Example results on MillionAID training images at different masking ratios from Table~\ref{tab_mask_ratio}. It can be seen that MA3E successfully models the basic structure of the scenes in the original images and restores the preset angle variations. Even with the rotated crop retaining only 7 visible patches (80\% masking), the model still exhibits excellent angle restoration ability. This demonstrates that MA3E has learned rotation-invariant representations and can infer complex reconstructions.}
\label{fig_vis_diff_ratios}
\end{figure}

\begin{table}[t]
\begin{minipage}{0.49\textwidth}
\centering
\begin{tabular}{llccc}
$r$ & $b$ & ft & det & seg \\
\hline
67\% & 67\%   & 95.72 & 77.56 & 62.50 \\
67\% & 75\%   & 95.70 & 77.81 & 62.66 \\
75\% & 75\% & \cellcolor{lightgreen}\textbf{95.77} & \cellcolor{lightgreen}\textbf{77.93} & \cellcolor{lightgreen}\textbf{62.74} \\
80\% & 75\%   & 95.62 & 77.65 & 62.48 \\
80\% & 80\%   & 95.41 & 77.31 & 62.28
\end{tabular}
\caption{Masking ratios. `$r$' for the rotated crop and `$b$' for the background.}
\label{tab_mask_ratio}
\end{minipage}
\hfill
\begin{minipage}{0.49\textwidth}
\centering
\begin{tabular}{lccc}
Type & ft & det & seg \\
\hline
masked & 94.89 & 77.11 & 62.33 \\
all & \cellcolor{lightgreen}\textbf{95.77} & \cellcolor{lightgreen}\textbf{77.93} & \cellcolor{lightgreen}\textbf{62.74} \\
 & & & \\
& & & \\
& & &
\end{tabular}
\caption{Reconstructing types of patches within the rotated crop.}
\label{tab_ot_rec}
\end{minipage}
\end{table}

\begin{table}[t]
\rowcolors{3}{lightgreen}{lightgreen}
\centering
\begin{tabular}{lccccccc}
Methods & Backbone & Dataset & Epoch & GPU H. / Ep. & ft & det & seg \\
\hline
MAE~\cite{he2022masked} & ViT-S & MillionAID & 300 & 0.6 & 93.54 & 72.01 & 57.86 \\
MA3E & ViT-S & MillionAID & 300 & 0.7 & \textbf{93.89} & \textbf{74.23} & \textbf{58.46}
\end{tabular}
\caption{Comparison of results using ViT-S as the encoder.}
\label{tab_scale}
\end{table}

\begin{table}[t]
\rowcolors{4}{lightgreen}{lightgreen}
\centering
\begin{tabular}{lccccccc}
Methods & Backbone & Dataset & Epoch & GPU H. / Ep. & ft & det & seg \\
\hline
SatMAE~\cite{cong2022satmae}  & ViT-B & MillionAID & 300 & 1.2 & 95.40 & 75.96 & 60.93 \\
ScaleMAE~\cite{reed2023scale} & ViT-B & MillionAID & 300 & 2.0 & \textbf{95.89} & 75.97 & 61.58 \\
MA3E  & ViT-B & MillionAID & 300 & 1.4 & 95.77 & \textbf{77.93} & \textbf{62.74}
\end{tabular}
\caption{Results of different methods pre-trained on MillionAID.}
\label{tab_fairness}
\end{table}

\section{Additional Ablation Results}
We conduct additional ablation studies, maintaining the same experimental settings as described in the main paper.

\begin{figure}[!t]
\centering
\includegraphics[width=4.8in]{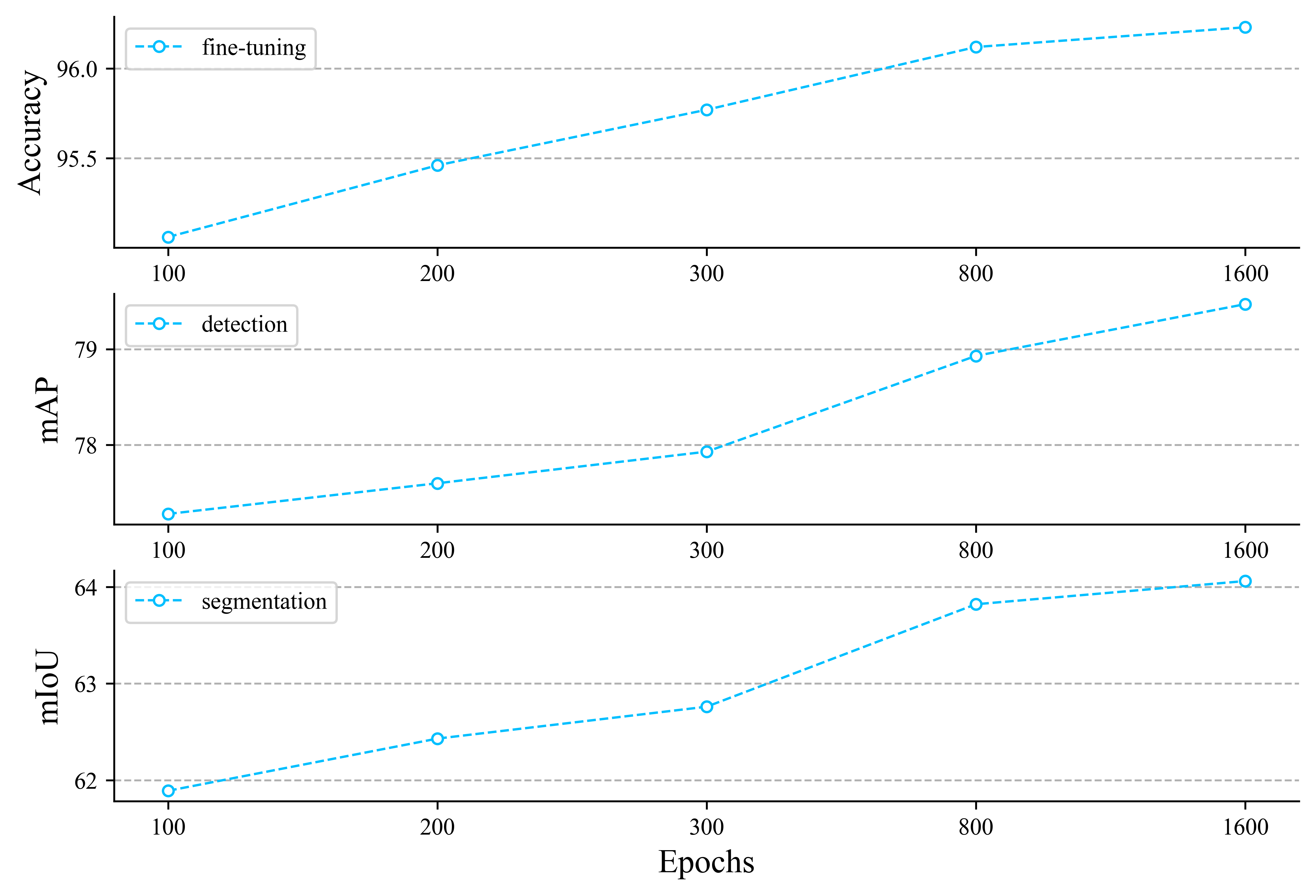}
\caption{The performance of MA3E on three tasks for different pre-training \textit{epochs}.}
\label{Fig_epoch}
\end{figure}

{\bf Masking ratio.} The results of randomly masking the rotated crop and the background with different ratios are in Table~\ref{tab_mask_ratio}. \cite{he2022masked} indicates that the optimal masking ratio is 75\%, so we explore suitable ratios around this value. It can be seen that our model achieves the best performance when the masking ratios of both the rotated crop and the background are 75\%. In addition, we visualize the reconstruction performance at different masking ratios in Fig.~\ref{fig_vis_diff_ratios}.

{\bf Reconstruction for rotated crops.} Table~\ref{tab_ot_rec} presents the ablation results of MA3E using ${\cal L}_{OT}$ to reconstruct only masked patches or reconstruct all patches within the rotated crop. The performance is better when reconstructing all patches because it also takes into account offset angles and scenes in the visible patches from the rotated crop.

{\bf The effect of pre-training \textit{epochs}.} Fig.~\ref{Fig_epoch} illustrates the performance of MA3E under different pre-training \textit{epochs}. The performance on three tasks improves gradually with the increase in \textit{epochs}, and MA3E pre-trained for 1600 \textit{epochs} has not reached saturation. It can be observed that longer training times may still have the potential for performance improvement, especially in rotated object detection.

{\bf Scalability.} Table~\ref{tab_scale} shows various downstream results of MA3E and MAE~\cite{he2022masked} when using ViT-S as the encoder. With only an extra 0.1 hours of training time per \textit{epoch}, MA3E achieves better results than MAE, showcasing its good scalability. Due to limited computational resources, we do not experiment with larger backbones such as ViT-L/H.

{\bf Unified Pre-training Dataset.} Considering the typical MIM methods for RS images, SatMAE~\cite{cong2022satmae} and ScaleMAE~\cite{reed2023scale}, use fMoW~\cite{christie2018functional} for pre-training, we follow their default settings and pre-train on MillionAID for a fair comparison. Note that MillionAID lacks multi-spectral bands, so we pre-train SatMAE according to it facing fMoW-RGB. In Table~\ref{tab_fairness}, MA3E achieves excellent downstream results with relative shorter training time per \textit{epoch}.

\section{Reconsidering OT}

\begin{figure}[!t]
\centering
\includegraphics[width=4.8in]{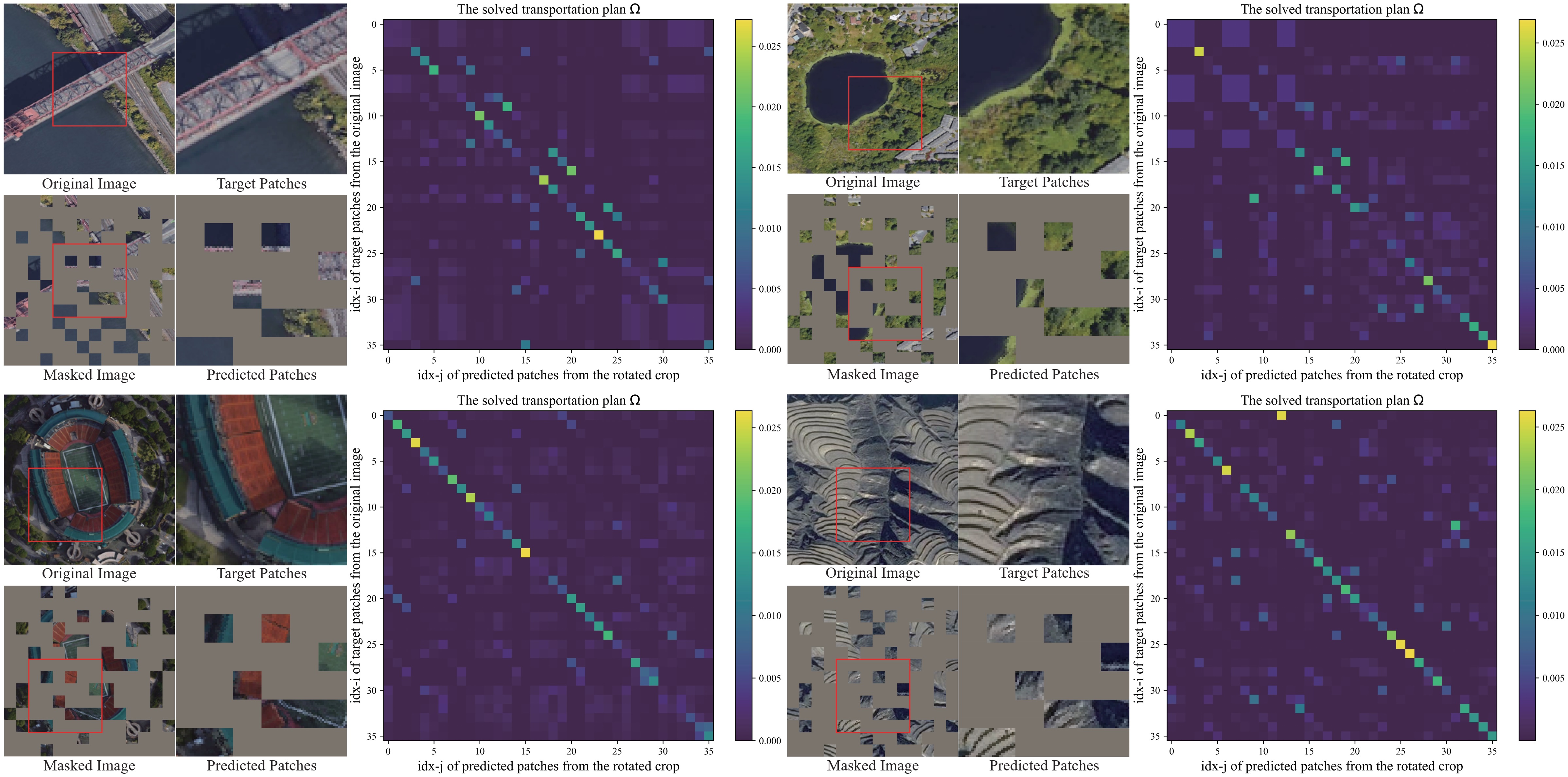}
\caption{Example heatmaps of the solved transportation plan $\Omega$ on MillionAID training images. For each set, the predicted patches is output from the decoder in practice, and we display masked images for ease of observation. After patchifying, indices are assigned to each target patch and predicted patch in a left-to-right, top-to-bottom order. The heatmap visualizes the weight calculated for the mean squared error between target patch at \texttt{idx\_i} and predicted patch at \texttt{idx\_j}. The weights sum up to $1/36$ for each column and total to 1 for the entire heatmap.}
\label{Fig_vis_ot}
\end{figure}

During the reconstruction of rotated crops, we already define original image patches as suppliers and predicted rotated crop patches as demanders, with the L2 similarity between patch pairs serving as the transportation cost. The transportation plan $\Omega$ of this OT problem can be solved via the Sinkhorn-Knopp Iteration~\cite{cuturi2013sinkhorn}, which transforms the complex marginal linear programming problem into a solution process over a smooth feasible domain by introducing an entropic regularization term. Note that this classic algorithm is textbook knowledge and not a contribution of this paper. For further details, please refer to prior works~\cite{cuturi2013sinkhorn, ge2021ota}.

MA3E leverages $\Omega = \{\omega_{i,j}\}$ to allocate similar original image patches as reconstruction targets for each predicted patch. Fig.~\ref{Fig_vis_ot} displays some heatmaps of the sovled transportation plan. Essentially, $\omega_{i,j}$ can be seen as the weight used in computing the mean squared error between the $i$-th target patch and $j$-th predicted patch in ${\cal L}_{OT}$, where higher weights are assigned to more similar target-prediction pairs. When reconstructing the $j$-th predicted patch, the model computes a weighted sum of MSE between it and multiple target patches ($j$-th column in the heatmap of Fig.~\ref{Fig_vis_ot}). It's evident that each predicted patch matches similar target patches distributed across multiple positions rather than just at the same location. This demonstrates the effectiveness of MA3E in reconstructing rotated crops by solving the OT problem.

\section{More Visualizations}

\begin{figure}[!htpb]
\centering
\includegraphics[width=4.8in]{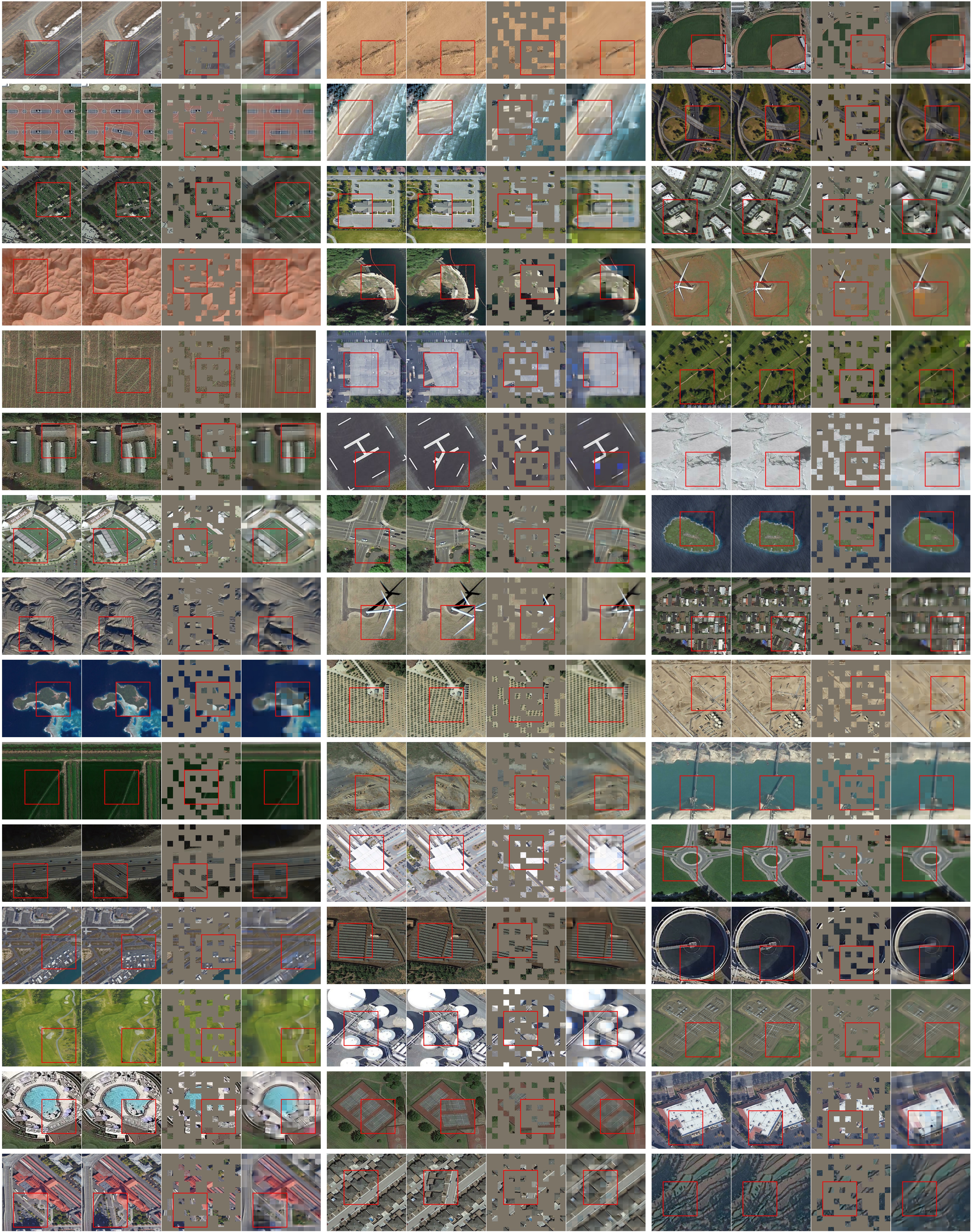}
\caption{More example results on MillionAID training images. For each set, from left to right, there are the original image, the composite image, the masked image, and the reconstructed image.}
\label{fig_vis_more}
\end{figure}

\begin{figure}[!htpb]
\centering
\includegraphics[width=4.8in]{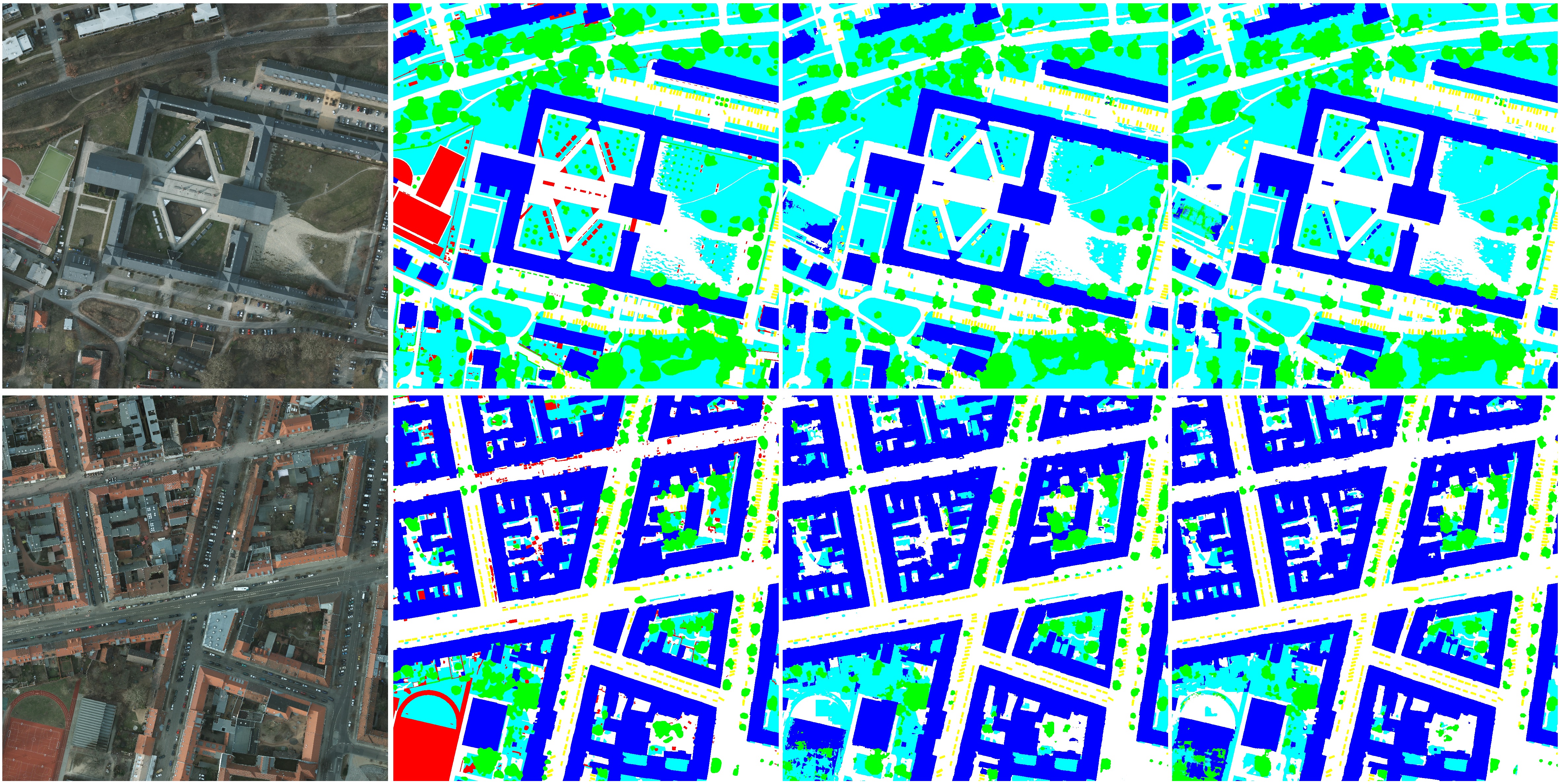}
\caption{Segmentation maps examples of fine-tuning different pre-trained models on Potsdam testing images. For each row, we sequentially display the original image, the ground truth, results from MA3E pre-trained for 1600 \textit{epochs}, and results from \textit{Wang et al.}~\cite{wang2022advancing}'s MAE pre-trained for 1600 \textit{epochs}. During evaluation, the clutter class (red in the ground truth) are often ignored. Compared to \textit{Wang et al.}, MA3E exhibits segmentation details that are richer and closer to the ground truth. Best viewed with zoom-in.}
\label{fig_seg}
\end{figure}

\begin{figure}[!htpb]
\centering
\includegraphics[width=4.8in]{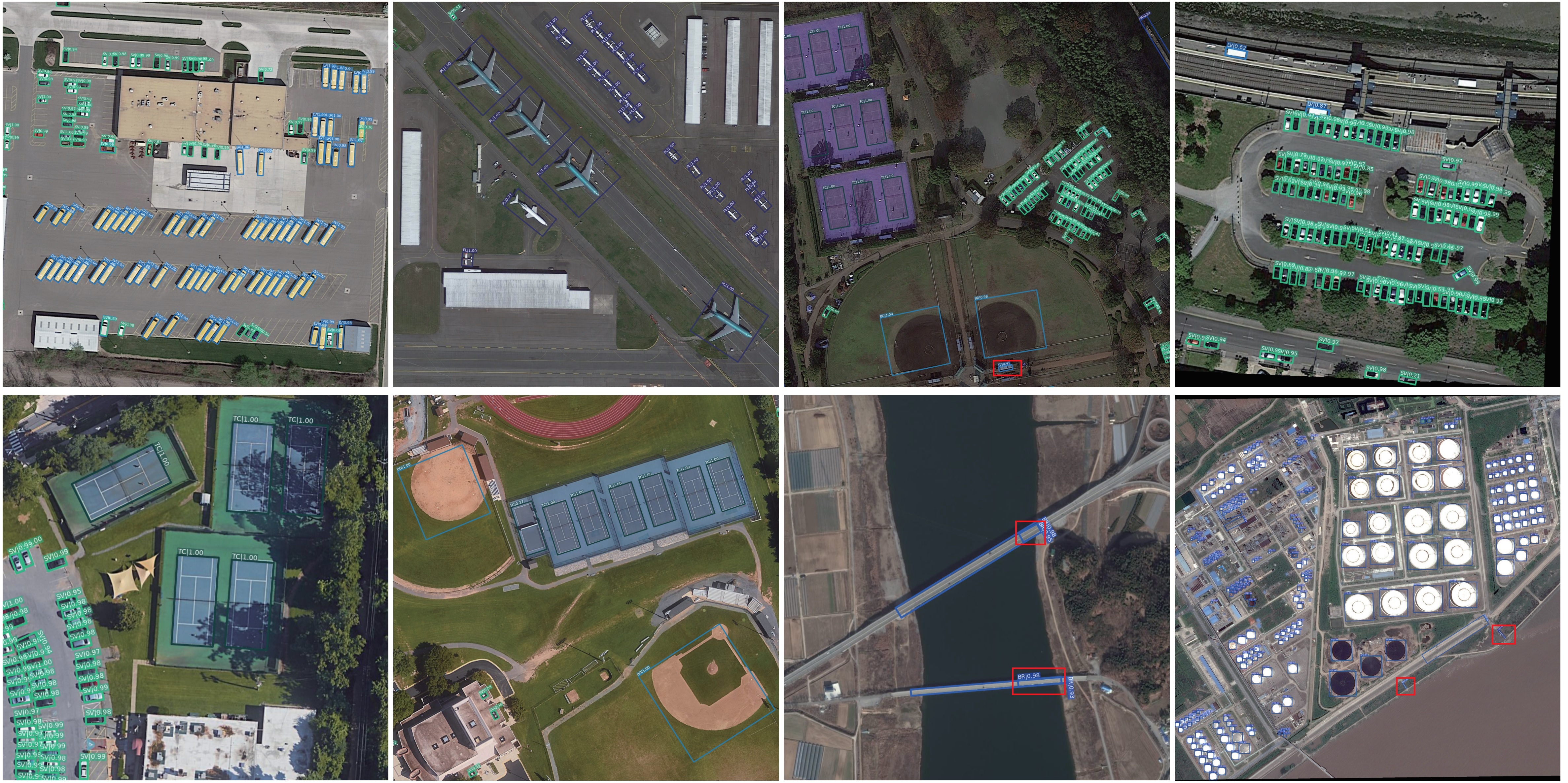}
\caption{Detection results examples of MA3E of 1600 \textit{epochs} on DOTA1.0 testing images. When fine-tuning MA3E, the detector effectively detects objects of various scales and dense distributions. However, it still encounters false positives when dealing with objects highlighted in red boxes, which have larger aspect ratios and complex backgrounds. Best viewed with zoom-in.}
\label{fig_det}
\end{figure}

In this section, we show more MA3E reconstructed images in Fig.~\ref{fig_vis_more}. Rotated crops are marked with red bounding boxes. For the downstream tasks, parts of semantic segmentation results and rotated object detection results are further presented in Fig.~\ref{fig_seg} and Fig.~\ref{fig_det}, respectively.

%
%

\begin{thebibliography}{10}
\providecommand{\url}[1]{\texttt{#1}}
\providecommand{\urlprefix}{URL }
\providecommand{\doi}[1]{https://doi.org/#1}

\bibitem{bao2021beit}
Bao, H., Dong, L., Piao, S., Wei, F.: Beit: Bert pre-training of image transformers. arXiv preprint arXiv:2106.08254  (2021)

\bibitem{caron2021emerging}
Caron, M., Touvron, H., Misra, I., J{\'e}gou, H., Mairal, J., Bojanowski, P., Joulin, A.: Emerging properties in self-supervised vision transformers. In: Proceedings of the IEEE/CVF international conference on computer vision. pp. 9650--9660 (2021)

\bibitem{chen2022efficient}
Chen, J., Hu, M., Li, B., Elhoseiny, M.: Efficient self-supervised vision pretraining with local masked reconstruction. arXiv preprint arXiv:2206.00790  (2022)

\bibitem{chen2020generative}
Chen, M., Radford, A., Child, R., Wu, J., Jun, H., Luan, D., Sutskever, I.: Generative pretraining from pixels. In: International conference on machine learning. pp. 1691--1703. PMLR (2020)

\bibitem{chen2020simple}
Chen, T., Kornblith, S., Norouzi, M., Hinton, G.: A simple framework for contrastive learning of visual representations. In: International conference on machine learning. pp. 1597--1607. PMLR (2020)

\bibitem{chen2021empirical}
Chen, X., Xie, S., He, K.: An empirical study of training self-supervised vision transformers. In: Proceedings of the IEEE/CVF International Conference on Computer Vision. pp. 9620--9629 (2021)

\bibitem{cheng2017remote}
Cheng, G., Han, J., Lu, X.: Remote sensing image scene classification: Benchmark and state of the art. Proceedings of the IEEE  \textbf{105}(10),  1865--1883 (2017)

\bibitem{cheng2022anchor}
Cheng, G., Wang, J., Li, K., Xie, X., Lang, C., Yao, Y., Han, J.: Anchor-free oriented proposal generator for object detection. IEEE Transactions on Geoscience and Remote Sensing  \textbf{60},  1--11 (2022)

\bibitem{christie2018functional}
Christie, G., Fendley, N., Wilson, J., Mukherjee, R.: Functional map of the world. In: Proceedings of the IEEE Conference on Computer Vision and Pattern Recognition. pp. 6172--6180 (2018)

\bibitem{clark2020electra}
Clark, K., Luong, M.T., Le, Q.V., Manning, C.D.: Electra: Pre-training text encoders as discriminators rather than generators. arXiv preprint arXiv:2003.10555  (2020)

\bibitem{cong2022satmae}
Cong, Y., Khanna, S., Meng, C., Liu, P., Rozi, E., He, Y., Burke, M., Lobell, D., Ermon, S.: Satmae: Pre-training transformers for temporal and multi-spectral satellite imagery. Advances in Neural Information Processing Systems  \textbf{35},  197--211 (2022)

\bibitem{contributors2020mmsegmentation}
Contributors, M.: Mmsegmentation: Openmmlab semantic segmentation toolbox and benchmark (2020)

\bibitem{cubuk2020randaugment}
Cubuk, E.D., Zoph, B., Shlens, J., Le, Q.V.: Randaugment: Practical automated data augmentation with a reduced search space. In: Proceedings of the IEEE/CVF conference on computer vision and pattern recognition workshops. pp. 702--703 (2020)

\bibitem{cuturi2013sinkhorn}
Cuturi, M.: Sinkhorn distances: Lightspeed computation of optimal transport. Advances in neural information processing systems  \textbf{26} (2013)

\bibitem{deng2009imagenet}
Deng, J., Dong, W., Socher, R., Li, L.J., Li, K., Fei-Fei, L.: Imagenet: A large-scale hierarchical image database. In: 2009 IEEE conference on computer vision and pattern recognition. pp. 248--255. Ieee (2009)

\bibitem{dosovitskiy2020image}
Dosovitskiy, A., Beyer, L., Kolesnikov, A., Weissenborn, D., Zhai, X., Unterthiner, T., Dehghani, M., Minderer, M., Heigold, G., Gelly, S., et~al.: An image is worth 16x16 words: Transformers for image recognition at scale. arXiv preprint arXiv:2010.11929  (2020)

\bibitem{drusch2012sentinel}
Drusch, M., Del~Bello, U., Carlier, S., Colin, O., Fernandez, V., Gascon, F., Hoersch, B., Isola, C., Laberinti, P., Martimort, P., et~al.: Sentinel-2: Esa's optical high-resolution mission for gmes operational services. Remote sensing of Environment  \textbf{120},  25--36 (2012)

\bibitem{gao2022convmae}
Gao, P., Ma, T., Li, H., Lin, Z., Dai, J., Qiao, Y.: Convmae: Masked convolution meets masked autoencoders. arXiv preprint arXiv:2205.03892  (2022)

\bibitem{gao2023mimic}
Gao, P., Zhang, R., Fang, R., Lin, Z., Li, H., Li, H., Yu, Q.: Mimic before reconstruct: Enhancing masked autoencoders with feature mimicking. arXiv preprint arXiv:2303.05475  (2023)

\bibitem{ge2021ota}
Ge, Z., Liu, S., Li, Z., Yoshie, O., Sun, J.: Ota: Optimal transport assignment for object detection. In: Proceedings of the IEEE/CVF Conference on Computer Vision and Pattern Recognition. pp. 303--312 (2021)

\bibitem{glorot2010understanding}
Glorot, X., Bengio, Y.: Understanding the difficulty of training deep feedforward neural networks. In: Proceedings of the thirteenth international conference on artificial intelligence and statistics. pp. 249--256. JMLR Workshop and Conference Proceedings (2010)

\bibitem{goyal2017accurate}
Goyal, P., Doll{\'a}r, P., Girshick, R., Noordhuis, P., Wesolowski, L., Kyrola, A., Tulloch, A., Jia, Y., He, K.: Accurate, large minibatch sgd: Training imagenet in 1 hour. arXiv preprint arXiv:1706.02677  (2017)

\bibitem{grill2020bootstrap}
Grill, J.B., Strub, F., Altch{\'e}, F., Tallec, C., Richemond, P., Buchatskaya, E., Doersch, C., Avila~Pires, B., Guo, Z., Gheshlaghi~Azar, M., et~al.: Bootstrap your own latent-a new approach to self-supervised learning. Advances in neural information processing systems  \textbf{33},  21271--21284 (2020)

\bibitem{gu2019survey}
Gu, Y., Wang, Y., Li, Y.: A survey on deep learning-driven remote sensing image scene understanding: Scene classification, scene retrieval and scene-guided object detection. Applied Sciences  \textbf{9}(10), ~2110 (2019)

\bibitem{he2022masked}
He, K., Chen, X., Xie, S., Li, Y., Doll{\'a}r, P., Girshick, R.: Masked autoencoders are scalable vision learners. In: Proceedings of the IEEE/CVF conference on computer vision and pattern recognition. pp. 16000--16009 (2022)

\bibitem{he2020momentum}
He, K., Fan, H., Wu, Y., Xie, S., Girshick, R.: Momentum contrast for unsupervised visual representation learning. In: Proceedings of the IEEE/CVF conference on computer vision and pattern recognition. pp. 9729--9738 (2020)

\bibitem{hou2022milan}
Hou, Z., Sun, F., Chen, Y.K., Xie, Y., Kung, S.Y.: Milan: Masked image pretraining on language assisted representation. arXiv preprint arXiv:2208.06049  (2022)

\bibitem{huang2016deep}
Huang, G., Sun, Y., Liu, Z., Sedra, D., Weinberger, K.Q.: Deep networks with stochastic depth. In: Computer Vision--ECCV 2016: 14th European Conference, Amsterdam, The Netherlands, October 11--14, 2016, Proceedings, Part IV 14. pp. 646--661. Springer (2016)

\bibitem{huang2022green}
Huang, L., You, S., Zheng, M., Wang, F., Qian, C., Yamasaki, T.: Green hierarchical vision transformer for masked image modeling. Advances in Neural Information Processing Systems  \textbf{35},  19997--20010 (2022)

\bibitem{ippoliti2019defining}
Ippoliti, C., Candeloro, L., Gilbert, M., Goffredo, M., Mancini, G., Curci, G., Falasca, S., Tora, S., Di~Lorenzo, A., Quaglia, M., et~al.: Defining ecological regions in italy based on a multivariate clustering approach: A first step towards a targeted vector borne disease surveillance. PloS one  \textbf{14}(7),  e0219072 (2019)

\bibitem{li2022uniform}
Li, X., Wang, W., Yang, L., Yang, J.: Uniform masking: Enabling mae pre-training for pyramid-based vision transformers with locality. arXiv preprint arXiv:2205.10063  (2022)

\bibitem{li2022exploring}
Li, Y., Mao, H., Girshick, R., He, K.: Exploring plain vision transformer backbones for object detection. In: European Conference on Computer Vision. pp. 280--296. Springer (2022)

\bibitem{lin2014microsoft}
Lin, T.Y., Maire, M., Belongie, S., Hays, J., Perona, P., Ramanan, D., Doll{\'a}r, P., Zitnick, C.L.: Microsoft coco: Common objects in context. In: Computer Vision--ECCV 2014: 13th European Conference, Zurich, Switzerland, September 6-12, 2014, Proceedings, Part V 13. pp. 740--755. Springer (2014)

\bibitem{liu2023mixmae}
Liu, J., Huang, X., Zheng, J., Liu, Y., Li, H.: Mixmae: Mixed and masked autoencoder for efficient pretraining of hierarchical vision transformers. In: Proceedings of the IEEE/CVF Conference on Computer Vision and Pattern Recognition. pp. 6252--6261 (2023)

\bibitem{liu2023pixmim}
Liu, Y., Zhang, S., Chen, J., Chen, K., Lin, D.: Pixmim: Rethinking pixel reconstruction in masked image modeling. arXiv preprint arXiv:2303.02416  (2023)

\bibitem{liu2023improving}
Liu, Y., Zhang, S., Chen, J., Yu, Z., Chen, K., Lin, D.: Improving pixel-based mim by reducing wasted modeling capability. In: Proceedings of the IEEE/CVF International Conference on Computer Vision. pp. 5361--5372 (2023)

\bibitem{liu2021swin}
Liu, Z., Lin, Y., Cao, Y., Hu, H., Wei, Y., Zhang, Z., Lin, S., Guo, B.: Swin transformer: Hierarchical vision transformer using shifted windows. In: Proceedings of the IEEE/CVF international conference on computer vision. pp. 10012--10022 (2021)

\bibitem{long2021creating}
Long, Y., Xia, G.S., Li, S., Yang, W., Yang, M.Y., Zhu, X.X., Zhang, L., Li, D.: On creating benchmark dataset for aerial image interpretation: Reviews, guidances, and million-aid. IEEE Journal of selected topics in applied earth observations and remote sensing  \textbf{14},  4205--4230 (2021)

\bibitem{loshchilov2016sgdr}
Loshchilov, I., Hutter, F.: Sgdr: Stochastic gradient descent with warm restarts. arXiv preprint arXiv:1608.03983  (2016)

\bibitem{loshchilov2017decoupled}
Loshchilov, I., Hutter, F.: Decoupled weight decay regularization. arXiv preprint arXiv:1711.05101  (2017)

\bibitem{mall2023change}
Mall, U., Hariharan, B., Bala, K.: Change-aware sampling and contrastive learning for satellite images. In: Proceedings of the IEEE/CVF Conference on Computer Vision and Pattern Recognition. pp. 5261--5270 (2023)

\bibitem{manas2021seasonal}
Manas, O., Lacoste, A., Gir{\'o}-i Nieto, X., Vazquez, D., Rodriguez, P.: Seasonal contrast: Unsupervised pre-training from uncurated remote sensing data. In: Proceedings of the IEEE/CVF International Conference on Computer Vision. pp. 9414--9423 (2021)

\bibitem{mendieta2023towards}
Mendieta, M., Han, B., Shi, X., Zhu, Y., Chen, C.: Towards geospatial foundation models via continual pretraining. In: Proceedings of the IEEE/CVF International Conference on Computer Vision. pp. 16806--16816 (2023)

\bibitem{muhtar2023cmid}
Muhtar, D., Zhang, X., Xiao, P., Li, Z., Gu, F.: Cmid: A unified self-supervised learning framework for remote sensing image understanding. IEEE Transactions on Geoscience and Remote Sensing  (2023)

\bibitem{mulla2013twenty}
Mulla, D.J.: Twenty five years of remote sensing in precision agriculture: Key advances and remaining knowledge gaps. Biosystems engineering  \textbf{114}(4),  358--371 (2013)

\bibitem{reed2023scale}
Reed, C.J., Gupta, R., Li, S., Brockman, S., Funk, C., Clipp, B., Keutzer, K., Candido, S., Uyttendaele, M., Darrell, T.: Scale-mae: A scale-aware masked autoencoder for multiscale geospatial representation learning. In: Proceedings of the IEEE/CVF International Conference on Computer Vision. pp. 4088--4099 (2023)

\bibitem{rolfe2016discrete}
Rolfe, J.T.: Discrete variational autoencoders. arXiv preprint arXiv:1609.02200  (2016)

\bibitem{rolnick2022tackling}
Rolnick, D., Donti, P.L., Kaack, L.H., Kochanski, K., Lacoste, A., Sankaran, K., Ross, A.S., Milojevic-Dupont, N., Jaques, N., Waldman-Brown, A., et~al.: Tackling climate change with machine learning. ACM Computing Surveys (CSUR)  \textbf{55}(2),  1--96 (2022)

\bibitem{schumann2018assisting}
Schumann, G.J., Brakenridge, G.R., Kettner, A.J., Kashif, R., Niebuhr, E.: Assisting flood disaster response with earth observation data and products: A critical assessment. Remote sensing  \textbf{10}(8), ~1230 (2018)

\bibitem{sun2022ringmo}
Sun, X., Wang, P., Lu, W., Zhu, Z., Lu, X., He, Q., Li, J., Rong, X., Yang, Z., Chang, H., et~al.: Ringmo: A remote sensing foundation model with masked image modeling. IEEE Transactions on Geoscience and Remote Sensing  (2022)

\bibitem{szegedy2016rethinking}
Szegedy, C., Vanhoucke, V., Ioffe, S., Shlens, J., Wojna, Z.: Rethinking the inception architecture for computer vision. In: Proceedings of the IEEE conference on computer vision and pattern recognition. pp. 2818--2826 (2016)

\bibitem{tian2022beyond}
Tian, Y., Xie, L., Fang, J., Shi, M., Peng, J., Zhang, X., Jiao, J., Tian, Q., Ye, Q.: Beyond masking: Demystifying token-based pre-training for vision transformers. arXiv preprint arXiv:2203.14313  (2022)

\bibitem{uijlings2013selective}
Uijlings, J.R., Van De~Sande, K.E., Gevers, T., Smeulders, A.W.: Selective search for object recognition. International journal of computer vision  \textbf{104},  154--171 (2013)

\bibitem{wang2022advancing}
Wang, D., Zhang, Q., Xu, Y., Zhang, J., Du, B., Tao, D., Zhang, L.: Advancing plain vision transformer toward remote sensing foundation model. IEEE Transactions on Geoscience and Remote Sensing  \textbf{61},  1--15 (2022)

\bibitem{wang2023masked}
Wang, H., Tang, Y., Wang, Y., Guo, J., Deng, Z.H., Han, K.: Masked image modeling with local multi-scale reconstruction. In: Proceedings of the IEEE/CVF Conference on Computer Vision and Pattern Recognition. pp. 2122--2131 (2023)

\bibitem{wang2021pyramid}
Wang, W., Xie, E., Li, X., Fan, D.P., Song, K., Liang, D., Lu, T., Luo, P., Shao, L.: Pyramid vision transformer: A versatile backbone for dense prediction without convolutions. In: Proceedings of the IEEE/CVF international conference on computer vision. pp. 568--578 (2021)

\bibitem{waqas2019isaid}
Waqas~Zamir, S., Arora, A., Gupta, A., Khan, S., Sun, G., Shahbaz~Khan, F., Zhu, F., Shao, L., Xia, G.S., Bai, X.: isaid: A large-scale dataset for instance segmentation in aerial images. In: Proceedings of the IEEE/CVF Conference on Computer Vision and Pattern Recognition Workshops. pp. 28--37 (2019)

\bibitem{wei2022masked}
Wei, C., Fan, H., Xie, S., Wu, C.Y., Yuille, A., Feichtenhofer, C.: Masked feature prediction for self-supervised visual pre-training. In: Proceedings of the IEEE/CVF Conference on Computer Vision and Pattern Recognition. pp. 14668--14678 (2022)

\bibitem{xia2018dota}
Xia, G.S., Bai, X., Ding, J., Zhu, Z., Belongie, S., Luo, J., Datcu, M., Pelillo, M., Zhang, L.: Dota: A large-scale dataset for object detection in aerial images. In: Proceedings of the IEEE conference on computer vision and pattern recognition. pp. 3974--3983 (2018)

\bibitem{xia2017aid}
Xia, G.S., Hu, J., Hu, F., Shi, B., Bai, X., Zhong, Y., Zhang, L., Lu, X.: Aid: A benchmark data set for performance evaluation of aerial scene classification. IEEE Transactions on Geoscience and Remote Sensing  \textbf{55}(7),  3965--3981 (2017)

\bibitem{xiao2018unified}
Xiao, T., Liu, Y., Zhou, B., Jiang, Y., Sun, J.: Unified perceptual parsing for scene understanding. In: Proceedings of the European conference on computer vision (ECCV). pp. 418--434 (2018)

\bibitem{xie2022masked}
Xie, J., Li, W., Zhan, X., Liu, Z., Ong, Y.S., Loy, C.C.: Masked frequency modeling for self-supervised visual pre-training. arXiv preprint arXiv:2206.07706  (2022)

\bibitem{xie2021oriented}
Xie, X., Cheng, G., Wang, J., Yao, X., Han, J.: Oriented r-cnn for object detection. In: Proceedings of the IEEE/CVF International Conference on Computer Vision. pp. 3520--3529 (2021)

\bibitem{xie2022simmim}
Xie, Z., Zhang, Z., Cao, Y., Lin, Y., Bao, J., Yao, Z., Dai, Q., Hu, H.: Simmim: A simple framework for masked image modeling. In: Proceedings of the IEEE/CVF Conference on Computer Vision and Pattern Recognition. pp. 9653--9663 (2022)

\bibitem{yang2010bag}
Yang, Y., Newsam, S.: Bag-of-visual-words and spatial extensions for land-use classification. In: Proceedings of the 18th SIGSPATIAL international conference on advances in geographic information systems. pp. 270--279 (2010)

\bibitem{you2017large}
You, Y., Gitman, I., Ginsburg, B.: Large batch training of convolutional networks. arXiv preprint arXiv:1708.03888  (2017)

\bibitem{yun2019cutmix}
Yun, S., Han, D., Oh, S.J., Chun, S., Choe, J., Yoo, Y.: Cutmix: Regularization strategy to train strong classifiers with localizable features. In: Proceedings of the IEEE/CVF international conference on computer vision. pp. 6023--6032 (2019)

\bibitem{zhang2017mixup}
Zhang, H., Cisse, M., Dauphin, Y.N., Lopez-Paz, D.: mixup: Beyond empirical risk minimization. arXiv preprint arXiv:1710.09412  (2017)

\bibitem{zhang2020domain}
Zhang, J., Liu, J., Pan, B., Shi, Z.: Domain adaptation based on correlation subspace dynamic distribution alignment for remote sensing image scene classification. IEEE Transactions on Geoscience and Remote Sensing  \textbf{58}(11),  7920--7930 (2020)

\bibitem{zhang2023vitaev2}
Zhang, Q., Xu, Y., Zhang, J., Tao, D.: Vitaev2: Vision transformer advanced by exploring inductive bias for image recognition and beyond. International Journal of Computer Vision pp. 1--22 (2023)

\bibitem{zhou2019semantic}
Zhou, B., Zhao, H., Puig, X., Xiao, T., Fidler, S., Barriuso, A., Torralba, A.: Semantic understanding of scenes through the ade20k dataset. International Journal of Computer Vision  \textbf{127},  302--321 (2019)

\end{thebibliography}


\end{document}